%% file: main.tex
\documentclass[10pt]{article}

\usepackage[margin=0.9in]{geometry}
\usepackage[T1]{fontenc}
\usepackage[utf8]{inputenc}
\usepackage{lmodern}
\usepackage{microtype}

\usepackage{amsmath,amssymb,amsthm}
\usepackage{bm}
\newtheorem{proposition}{Proposition}
\newtheorem{corollary}{Corollary}

\usepackage{booktabs}
\usepackage{multirow}
\usepackage{tabularx}
\usepackage{array}
\usepackage{graphicx}
\graphicspath{{figures/}{../assets/paper1_figs/}{assets/paper1_figs/}}
\usepackage{float}
\usepackage{placeins}
\usepackage{lscape}
\makeatletter
\g@addto@macro\landscape{\clearpage\pdfpageattr{/Rotate 90}}
\g@addto@macro\endlandscape{\clearpage\pdfpageattr{}}
\makeatother
\usepackage[font=small,labelfont=bf,skip=4pt]{caption}
\usepackage{subcaption}

\usepackage{xcolor}
\usepackage{listings}
\usepackage[hidelinks]{hyperref}
\usepackage{url}
\usepackage{cleveref}

\usepackage{enumitem}
\setlist{nosep,leftmargin=*}
\usepackage[normalem]{ulem}
\usepackage{seqsplit}

\newcommand{\code}[1]{\texttt{#1}}
\newcommand{\codebrk}[1]{\texttt{\small\seqsplit{#1}}}
\newcommand{\stdmax}{\ensuremath{\sigma_{\max}}}

\ifdefined\blindsubmission
  \newcommand{\paperauthors}{Anonymous Authors}
\else
  \input{arxiv_metadata.tex}

  \newcommand{\paperauthors}{\arxivauthors}
\fi

\title{\textbf{Diagnosing JEPA World Models with\\Action-Conditioned Predictive Consistency}}
\author{\paperauthors}
\date{}
\hypersetup{
  pdftitle={Diagnosing JEPA World Models with Action-Conditioned Predictive Consistency},
  pdfkeywords={JEPA world models, predictive consistency, action-conditioned rollouts, visual invariance, visual robustness}
}

\begin{document}
\maketitle
\ifdefined\blindsubmission\else
  \begingroup
  \renewcommand{\thefootnote}{}
  \makeatletter
  \def\Hy@footnote@currentHref{paper.authornotes}
  \makeatother
  \footnotetext{\arxivauthornotes}
  \endgroup
\fi

\begin{abstract}
Joint-embedding predictive architectures (JEPAs) learn world models that
predict in a compact latent space rather than in pixels, reducing the
pressure to model nuisance appearance. Yet this provides no guarantee against visual perturbations: they can
still alter the encoded representation and affect subsequent
action-conditioned predictions.
Bisimulation captures this requirement precisely: two observations should be
treated as the same state only when their action-conditioned consequences
agree. Guided by this criterion, we introduce Action-Conditioned Predictive
Consistency (ACPC), a diagnostic that measures how far a clean history and
a visually perturbed view of it diverge after being rolled forward under the
same action sequence. We prove that this divergence bounds the
perturbation-induced change in multi-step prediction error and planner cost.
Building on pairwise ACPC, we define two complementary measures: the
Invariance Radius (IR) summarizes clean--perturbed rollout spread, while the
Separation Rate (SR) checks whether different states remain distinguishable
after rollout.
Experiments on four visual control tasks show that pairwise ACPC predicts
perturbation-induced prediction and cost changes. On LeWM, the IR--SR screen
transfers across tasks, and the joint diagnostic remains
informative under blur and resize. PLDM exhibits similar diagnostic trends
under a different architecture.\ifdefined\blindsubmission\else\
Code is available \href{\paperpubliccodeurl}{\textcolor{blue}{here}}.\fi

\end{abstract}

\smallskip
\ifdefined\blindsubmission
\noindent\textbf{Anonymous reproducibility note:} code and data availability follow the venue's double-blind supplemental policy.
\fi

%% ===========================================================================
\section{Introduction}\label{sec:intro}

World models support control by predicting the outcomes of candidate
actions~\cite{hafner2025dreamerv3,hansen2024tdmpc2}. Joint-embedding
predictive architectures (JEPAs) predict target representations rather than
reconstructing pixels, avoiding an explicit requirement to reproduce nuisance
visual details
~\cite{lecun2022path,assran2023ijepa,bardes2024vjepa,vanassel2025jointembeddingreconstruction,littwin2024jepaavoidsnoisyfeatures}.
Yet this design alone does not determine which visual differences matter for
control.
Such representations should be insensitive to task-irrelevant visual
perturbations while preserving distinctions between situations that require
different actions.
This requirement echoes the intuition behind bisimulation, which defines
state equivalence through action-conditioned consequences rather than visual
appearance~\cite{gelada2019deepmdp,zhang2021dbc}. However, encoder distances
alone do not show how a perturbation propagates through prediction. Over a
multi-step rollout, the predictor may amplify or contract the initial
representation difference. Encoder-level and single-transition diagnostics
do not directly measure this rollout-level effect. We therefore ask: when a
clean history and its perturbed view are rolled forward under the same action
sequence, how far apart are their predicted trajectories?

\begin{figure}[t]
\centering
\includegraphics[width=\linewidth]{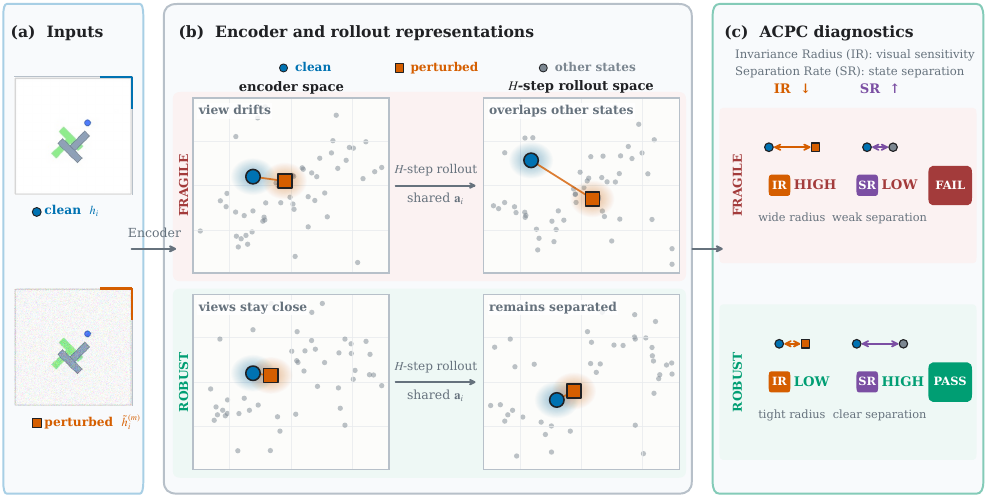}
\caption{\textbf{Conceptual overview of ACPC, IR, and SR.}
\textbf{(a)}~A logged clean history and a visually perturbed view of it form
the paired inputs.
\textbf{(b)}~A frozen world model encodes both views and rolls them forward
for $H$ steps under the same recorded action sequence. ACPC measures the
distance between the two predicted trajectories. In the fragile case,
perturbation-induced spread grows and overlaps representations of other
states (gray); in the robust case, the paired views remain close while other
states stay separated. Layouts are schematic: only within-panel proximity is
meaningful, and positions enter no reported metric.
\textbf{(c)}~The Invariance Radius (IR) summarizes perturbation-induced
rollout spread across histories, while the Separation Rate (SR) measures how
often selected different-state rollouts remain separated beyond this spread.
A checkpoint passes the diagnostic screen only if IR is low and SR is high;
passing is specific to the evaluated visual shift and state labels and does
not certify robustness.}
\label{fig:acpc-ir-sr-overview}
\end{figure}

To measure this rollout-level effect, we introduce Action-Conditioned
Predictive Consistency (ACPC). ACPC rolls a clean history and its perturbed
view forward under the same action sequence and measures the distance between
their predicted trajectories. \Cref{fig:acpc-ir-sr-overview} provides a
conceptual overview of how this pairwise distance is summarized by IR and
complemented by SR. Low ACPC alone, however, does not rule out
representational collapse: a constant representation would make every paired
rollout identical. We therefore summarize ACPC across histories with the
\emph{Invariance Radius} (IR), where lower values indicate less sensitivity
to visual perturbations. The \emph{Separation Rate} (SR) checks whether
different states remain distinguishable after rollout, with higher values
indicating better separation.

\FloatBarrier

We prove that ACPC bounds the perturbation-induced change in multi-step
prediction error. When the same candidate action sequences are evaluated
from both views, ACPC also yields bounds on changes in their predicted costs
(\Cref{prop:target-free-error-drift,prop:exact-cost-certificate}). These
bounds hold for each evaluated pair and require no distributional or
smoothness assumptions. Experiments on LeWM~\cite{maes2026lewm} show that
ACPC is informative about prediction-error change and the extra predicted
cost incurred when a visual perturbation causes the cross-entropy method
(CEM)~\cite{kroese2006cem} to select a different plan. Together, IR and SR
help identify checkpoints that perform well under visual perturbations across
tasks and perturbation types. On PLDM~\cite{sobal2025stresstesting}, better
planning performance under visual perturbation is also generally accompanied
by lower IR and higher SR, showing the same qualitative diagnostic pattern
under a second world-model architecture.

The contributions are:
\begin{enumerate}[leftmargin=1.4em]
  \item We introduce \textbf{Action-Conditioned Predictive Consistency
  (ACPC)}, which compares the predicted trajectories of a clean history and
  its perturbed view under the same action sequence. IR summarizes
  sensitivity across histories, while SR checks whether different states
  remain distinguishable (\Cref{sec:acpc}).
  \item We prove \textbf{samplewise bounds} on how much a visual perturbation
  can change multi-step prediction error. When the same candidate action
  sequences are evaluated from both views, we also bound changes in their
  predicted costs
  (\Cref{prop:target-free-error-drift,prop:exact-cost-certificate}).
  \item We evaluate \textbf{ACPC, IR, and SR} across four control tasks and
  three visual perturbations, and examine their behavior on a second
  world-model architecture. ACPC provides predictive information about
  perturbation-induced changes in multi-step prediction error and planning
  cost, while IR and SR help identify trained models that maintain control
  performance (\Cref{sec:exp}).
\end{enumerate}
\section{Related Work}\label{sec:related}

\paragraph{World Models.}
World models predict how an environment evolves under actions and use those
predictions for control. DreamerV3~\cite{hafner2025dreamerv3} learns from
imagined trajectories, while TD-MPC2~\cite{hansen2024tdmpc2} plans with
learned latent dynamics. Joint-embedding predictive architectures (JEPAs)
learn representations by predicting targets in representation space rather
than reconstructing pixels
~\cite{lecun2022path,assran2023ijepa,bardes2024vjepa,assran2025vjepa2}.
LeWM~\cite{maes2026lewm}, our primary model family, combines this
joint-embedding objective with action-conditioned latent prediction.
PLDM~\cite{sobal2025stresstesting}
provides a different joint-embedding dynamics architecture on which we also
evaluate ACPC. Earlier work analyzes how joint-embedding predictive objectives
can favor slow features~\cite{sobal2022jointembeddingpredictivearchitectures}.
Related training objectives predict future latent observations from histories
and actions in PBL~\cite{guo2020pbl}, or match predicted and target
representations under augmentation in SPR~\cite{schwarzer2021spr}.
Subsequent work studies collapse in self-predictive
learning~\cite{tang2023selfpredictive} and the role of action
conditioning~\cite{khetarpal2025actionconditional}. Theoretical analyses ask
when latent prediction can disregard irrelevant features or favor predictable,
high-influence features
~\cite{vanassel2025jointembeddingreconstruction,littwin2024jepaavoidsnoisyfeatures}
and when it can recover latent state up to a linear
transformation~\cite{klindt2026lejepaworldmodel}.
seq-JEPA~\cite{ghaemi2025seqjepa} further studies invariance and equivariance
under known view transformations. These works explain how predictive
representations and dynamics are learned; we evaluate how a trained predictor
responds when its visual input is perturbed.

\paragraph{Robustness to Visual Perturbations.}
Training-time augmentation improves visual control in
DrQ~\cite{kostrikov2020drq} and DrQ-v2~\cite{yarats2022drqv2}, while
SODA~\cite{hansen2021soda} uses augmentation in an auxiliary
representation-learning objective. ViGMO~\cite{vigmo} and
VIBR~\cite{dupuis2023vibr} learn invariance in latent or value space.
ReOI~\cite{chen2025reoi} instead modifies observations at test time to reduce
distractor sensitivity in visual model-predictive control. Other methods
change world-model training. TPC~\cite{nguyen2021tpc} favors temporally
predictable information, and DreamerPro~\cite{deng2022dreamerpro} replaces
pixel reconstruction with prototype prediction. Task Informed
Abstractions~\cite{fu2021tia} and Denoised
MDPs~\cite{wang2022denoisedmdps} use task or reward information to separate
useful state from distractors. Iso-Dream~\cite{pan2022isodream} separates
controllable and noncontrollable dynamics, while other approaches combine
temporal masking with bisimulation~\cite{sun2024latentrobustwm}, infer the
actions of visually similar distractors~\cite{wang2024ad3}, or use task-aware
reconstruction~\cite{hutson2024policyshaped}. These methods reflect a broader
principle: robustness should remove irrelevant visual variation without
discarding distinctions needed for control. Bisimulation formalizes
behavioral similarity through rewards and action-conditioned transitions
~\cite{gelada2019deepmdp,zhang2021dbc,bsmpc}, including a reward-free
formulation based only on transitions~\cite{toso2026bisimjepa}.
Value-equivalent models and value-aware model-learning objectives instead
focus on information needed for value prediction or
planning~\cite{grimm2020valueequivalence,voelcker2025calibratedvalueaware}.
Most closely related, MWM~\cite{yan2026mwm} enforces action-conditioned
rollout consistency during training. ACPC measures how clean and perturbed
latent rollouts diverge in a frozen model under the same action sequence,
without retraining it.

\paragraph{Diagnostics for World Models.}
Model accuracy and control performance do not always move together:
one-step likelihood can be a poor guide to downstream
performance~\cite{lambert2020objective}. This mismatch has motivated methods
that connect model learning and evaluation to downstream
decisions~\cite{wei2024objectivemismatch}. World-model evaluation should
therefore reflect the intended use of the model, including long rollouts,
planning, and policy evaluation~\cite{yu2026worldmodelevaluation}. One group
of methods tests or enforces whether learned dynamics respect actions.
ATM~\cite{chen2026atm} compares action information in encoded and predicted
transitions, while Delta-JEPA~\cite{zhang2026deltajepa} decodes actions from
latent differences. ACID~\cite{seo2026acid} adds inverse-dynamics cycle
consistency to the planning cost. Future-Compatible~\cite{ruan2026futurecompatible}
checks whether generated futures agree with their stated actions, and World
Models as Group Actions~\cite{wang2026groupactions} tests identity, inverse,
and composition structure. Related diagnostics expose kinematic failures in
long rollouts~\cite{schaefer2026kinematic} or use frozen inverse-dynamics
probes to test whether latents retain action information under visual
corruptions~\cite{yeom2026actionrelevant}. A second group evaluates downstream
failure. \mbox{WMAttack}~\cite{guo2026wmattack} searches for visual attacks on
closed-loop world-model agents, while ARB4WM~\cite{zhang2026arb4wm} targets
their policy, value, and latent dynamics. Other work compares predicted
multi-step latent transitions with those produced by the
environment~\cite{vakalis2026operator}, or compares predicted and true plan
costs~\cite{you2026controlpredictability}. Finally,
CARRL~\cite{lutjens2020certifiedrl} certifies Q-based action choices under
bounded observation uncertainty, and CROP~\cite{wu2022crop} certifies
per-state actions and finite-horizon reward bounds through smoothing. These
certificates cover policy decisions or return, whereas our bounds preserve
only the winner or elite set for an evaluated pair and fixed candidate pool.
More broadly, existing diagnostics assess action content, rollout validity,
model error, or attack sensitivity. ACPC asks how a task-preserving visual
perturbation propagates through a shared-action rollout without requiring the
true future.

\clearpage
\section{Action-Conditioned Predictive Consistency as a Diagnostic}\label{sec:acpc}

A visual perturbation can change the trajectory predicted by a world model.
ACPC measures this change by rolling a clean history and its perturbed version
forward under the same action sequence and comparing the two trajectories. We
first define this pairwise measurement. We then
show that it bounds perturbation-induced changes in prediction error and
candidate cost, which leads to conditions for preserving a planner's selected
candidate. To summarize ACPC across a checkpoint, IR captures
clean--perturbed pairs with high sensitivity, while SR checks that different
states remain separated and prevents a collapsed representation from
appearing robust. Finally, we combine IR and SR into an empirical checkpoint
screen.

\subsection{Pairwise ACPC}\label{sec:acpc-def}

Given a clean history and its perturbed version, ACPC compares the rollouts
predicted from them under identical actions. Let $h$ denote the clean history
and let $\tilde h$ be the result of applying the evaluated visual perturbation
to $h$. The frozen encoder maps them to $z=E_\theta(h)$ and
$\tilde z=E_\theta(\tilde h)$. Both representations are then rolled forward
under the action sequence $\mathbf a=(a_0,\ldots,a_{H-1})$. Let $F_\theta$
denote the frozen action-conditioned predictor. After the first $k$ actions,
the two predicted representations are
\begin{equation}
\hat z_k=F_\theta^k(z,\mathbf a_{0:k-1}),
\qquad
\hat{\tilde z}_k=F_\theta^k(\tilde z,\mathbf a_{0:k-1}).
\label{eq:acpc-rollout-objects}
\end{equation}
Here $H$ is the number of autoregressive future steps. ACPC compares
predictions in the projected latent space used to evaluate planner costs. Let
$\Pi$ map each predicted representation into this space. To combine all $H$
predicted steps into one trajectory-level distance, we assign nonnegative
weights $\alpha_k$, with $\sum_{k=1}^H\alpha_k=1$, and define the weighted
rollout vector
\begin{equation}
\bar G_{\mathbf a}(z)
=
\begin{bmatrix}
\sqrt{\alpha_1}\Pi(\hat z_1)^\top & \cdots &
\sqrt{\alpha_H}\Pi(\hat z_H)^\top
\end{bmatrix}^{\!\top}.
\label{eq:weighted-rollout-map}
\end{equation}
Action-Conditioned Predictive Consistency (ACPC) is the distance between the
two weighted predicted rollouts:
\begin{equation}
\mathrm{ACPC}_H(h,\tilde h,\mathbf a)
=
\left\|\bar G_{\mathbf a}(E_\theta(h))
-\bar G_{\mathbf a}(E_\theta(\tilde h))\right\|_2
=
\left(
\sum_{k=1}^H\alpha_k
\left\|\Pi(\hat z_k)-\Pi(\hat{\tilde z}_k)\right\|_2^2
\right)^{1/2}.
\label{eq:acpc-h}
\end{equation}
Low ACPC means that the two predicted trajectories remain close; high ACPC
means that they separate. Encoder shift $\|z-\tilde z\|_2$ measures the
difference before prediction, whereas ACPC measures it after rollout. Using
the same action sequence removes action differences from the comparison.
ACPC measures rollout consistency, not prediction accuracy.

We use uniform weights $\alpha_k=1/H$ and $H=8$, the longest horizon available
for every logged window (\Cref{sec:appendix-A}). The planner analysis uses
$H=5$ to match the CEM planning horizon (\Cref{sec:exp-planner}).

\subsection{Prediction-Error Bounds}

ACPC itself does not require the observed future. To connect it to prediction
error, let $Y_{\mathbf a}^H$ denote the observed future under the same action
sequence, represented in the same weighted space as the predicted rollouts
(\Cref{eq:weighted-rollout-map}). Since the perturbed history is created by
applying a visual perturbation to the clean history, both predicted rollouts
are compared with the same observed future. Their difference in error cannot
exceed the distance between the two predictions, which is ACPC.

\begin{proposition}[Prediction-error change bound]\label{prop:target-free-error-drift}
Define
\[
e_h=\|\bar G_{\mathbf a}(E_\theta(h))-Y_{\mathbf a}^H\|_2,
\qquad
e_{\tilde h}=\|\bar G_{\mathbf a}(E_\theta(\tilde h))-Y_{\mathbf a}^H\|_2.
\]
Then, for every paired sample,
\begin{equation}
|e_{\tilde h}-e_h|
\le \mathrm{ACPC}_H(h,\tilde h,\mathbf a).
\label{eq:target-free-error-drift}
\end{equation}
In particular, the increase in error caused by the perturbation,
$(e_{\tilde h}-e_h)_+$, is also bounded by ACPC.
\end{proposition}
This result follows directly from the reverse triangle inequality; the proof
is given in Appendix~\ref{sec:appendix-acpc-proofs}. The bound concerns the
change in error, not absolute prediction accuracy. Both rollouts can be
inaccurate even when ACPC is small. In the experiment, the observed future is
used to compute the error change $d=|e_{\tilde h}-e_h|$, not ACPC itself
(\Cref{sec:exp-target-aligned}).

\subsection{Selection Stability from Planning-Cost Bounds}

A planner ranks candidate action sequences by their predicted costs. Changing
the input can change each candidate's predicted endpoint and therefore its
cost. If these cost changes are too small to close the gaps between
candidates, the selected candidate remains unchanged. We now formalize this
idea.

For each candidate action sequence $\mathbf a^j$, we roll both inputs forward
under that sequence. Let
\[
x_j=\Pi(F_\theta^H(E_\theta(h),\mathbf a^j)),
\qquad
\tilde x_j=\Pi(F_\theta^H(E_\theta(\tilde h),\mathbf a^j))
\]
be the final clean and perturbed predicted representations. Their displacement
$r_j=\|x_j-\tilde x_j\|_2$ is one component of the candidate-specific ACPC.
Whenever $\alpha_H>0$,
\[
r_j\leq
\frac{\mathrm{ACPC}_H(h,\tilde h,\mathbf a^j)}{\sqrt{\alpha_H}}.
\]

\begin{proposition}[Planning-cost bounds]\label{prop:exact-cost-certificate}
Suppose the planner scores candidate $j$ by its squared distance to a fixed
goal embedding $g$:
\[
C_j=\|x_j-g\|_2^2,
\qquad
\tilde C_j=\|\tilde x_j-g\|_2^2.
\]
Define the candidate-specific cost-change bound
\begin{equation}
b_j=r_j\bigl(\|x_j-g\|_2+\|\tilde x_j-g\|_2\bigr).
\label{eq:exact-cost-bound}
\end{equation}
Then $|\tilde C_j-C_j|\leq b_j$.

Let $w$ and $\tilde w$ be the winners under the clean and perturbed inputs. If
the winner changes, the perturbed winner's excess cost under the clean
prediction satisfies
\begin{equation}
0\leq C_{\tilde w}-C_w\leq b_{\tilde w}+b_w.
\label{eq:fixed-pool-regret-bound}
\end{equation}
\end{proposition}

The bound $b_j$ is computed from the evaluated endpoints and requires no
global Lipschitz constant. If the clean winner $w$ leads every competitor $j$
by more than $b_w+b_j$, the perturbation cannot reverse their order. The same
argument preserves the top-$k$ elite set when every clean
elite--non-elite gap exceeds the corresponding pair of bounds.
Appendix~\ref{sec:appendix-acpc-proofs} gives the exact conditions
(\Cref{prop:fixed-pool-certificates}).

Because $r_j$ is bounded by candidate-specific ACPC, substituting its bound
into $b_j$ gives
\begin{equation}
|\tilde C_j-C_j|
\;\le\; b_j
\;\le\;
\frac{\mathrm{ACPC}_H(h,\tilde h,\mathbf a^j)}{\sqrt{\alpha_H}}
\bigl(\|x_j-g\|_2+\|\tilde x_j-g\|_2\bigr),
\label{eq:acpc-cost-chain}
\end{equation}
Thus, ACPC bounds how much each candidate's cost can change, while the clean
cost gaps determine whether that change can alter the planner's selection.
Checking these bounds requires evaluating every candidate under both inputs.
We therefore use them to analyze planner behavior after the runs are complete.

CEM plans in rounds. In each round, it scores a set of candidate action
sequences, keeps the best group (the elite set), and uses that group to
generate the candidates for the next round~\cite{kroese2006cem}. We compare a
clean and perturbed run that start from the same proposal and use the same
random samples. If the perturbation does not change the elite set in any
round, both runs generate the same candidates in the next round and
eventually return the same action
(\Cref{cor:adaptive-cem-alignment}). If the elite set changes, the guarantee
no longer applies. This result concerns the planner's model-based choice, not
its return in the environment.

\subsection{Checkpoint-Level IR and SR}\label{sec:acpc-ir-sr}

Pairwise ACPC measures one clean--perturbed pair under one action sequence.
Evaluating a checkpoint requires summarizing this measurement across many
histories. The \emph{Invariance Radius} (IR) measures how sensitive these
paired rollouts are to the visual perturbation; lower IR is better. Low IR
alone is not enough because a collapsed representation would make every
rollout look similar. The \emph{Separation Rate} (SR) therefore checks whether
different states remain distinguishable after rollout; higher SR is better.
A favorable checkpoint should have both low IR and high SR.

To compute IR, we select $n$ logged history windows as anchors. Anchor $i$
provides a clean history $h_i$ and its recorded action sequence
$\mathbf a_i=(a_{i,0},\ldots,a_{i,H-1})$. We apply the visual perturbation $M$
times to obtain $\tilde h_i^{(1)},\ldots,\tilde h_i^{(M)}$, and compute ACPC
for each clean--perturbed pair under the same recorded actions.

Raw latent distances can have different scales across histories. We therefore
divide each ACPC value by the anchor's typical one-step clean motion, denoted
by $s_i$. Starting from the final clean history frame, we measure the
projected displacement at each of the next $H$ observed transitions and take
the median
(\Cref{eq:clean-transition-scale,sec:appendix-A}). This expresses ACPC
relative to the amount of motion normally seen in that history. With a small
numerical stabilizer $\varepsilon_{\mathrm{norm}}$, define
\begin{equation}
R_i^{(m)}
=
\frac{
\mathrm{ACPC}_H(h_i,\tilde h_i^{(m)},\mathbf a_i)
}{s_i+\varepsilon_{\mathrm{norm}}},
\qquad
\bar R_i=\frac{1}{M}\sum_{m=1}^{M}R_i^{(m)}.
\label{eq:normalized-same-state-radius}
\end{equation}
The average $\bar R_i$ gives one normalized sensitivity value for each anchor.
Let $Q_q$ denote the empirical $q$-quantile across anchors. The raw IR is
\begin{equation}
\mathrm{IR}^{\mathrm{raw}}_q(\theta)
=Q_q\!\left(\{\bar R_i\}_{i=1}^{n}\right).
\label{eq:ir-raw}
\end{equation}
IR is one value for the checkpoint. We use $q=0.90$ so that IR reflects
anchors with relatively high sensitivity, which a mean could hide. The result
is stable for $q$ between $0.80$ and $0.95$ and for the tested rollout
horizons (\Cref{tab:horizon-quantile-sensitivity}).

IR checks whether perturbed views remain close to their clean counterparts. SR
checks whether this invariance also preserves differences between states. For
each eligible anchor, the protocol selects a nearby logged history with a
different state-coordinate label. Both histories are rolled forward under the
anchor's recorded actions, so their separation is not caused by different
action sequences. Their trajectory distance is normalized by the same
clean-motion scale used for IR and is denoted by $D_i^{\mathrm{diff}}$
(\Cref{eq:different-state-distance,sec:appendix-A}).

Let $\mathcal I$ be the set of anchors for which such a comparison is
available. SR is the fraction whose different-state distance exceeds raw IR
by a margin $\delta$:
\begin{equation}
\mathrm{SR}_{q,\delta}(\theta)
=
\frac{1}{|\mathcal I|}
\sum_{i\in\mathcal I}
\mathbf 1\!\left[
D_i^{\mathrm{diff}}
>
\mathrm{IR}^{\mathrm{raw}}_q(\theta)+\delta
\right],
\label{eq:sr}
\end{equation}
We use $q=0.90$ and $\delta=0.10$. SR therefore reports how often a tested
different-label pair remains beyond the checkpoint's perturbation radius plus
the fixed margin.

A collapsed representation makes IR small, but it also makes every
$D_i^{\mathrm{diff}}$ zero and therefore fails SR. Conversely, a checkpoint
can have high SR while remaining sensitive to visual perturbations, so IR is
still needed. Choosing these pairs requires state labels from the dataset. SR
therefore applies only to the labels used in the tested pairs.
Appendix~\ref{sec:appendix-A} gives the exact pairing rules and eligible-pair
counts.

\subsection{Checkpoint Screening}\label{sec:checkpoint-threshold-selection}

SR uses raw IR because it compares distances within the same checkpoint. To
compare a trained checkpoint with its unaugmented reference, we use relative
IR. Let $\theta_0$ denote the unaugmented checkpoint from the same task,
training run, and model family. We define
\begin{equation}
\mathrm{IR}^{\mathrm{rel}}_{q}(\theta;\theta_0)
=
\frac{\mathrm{IR}^{\mathrm{raw}}_{q}(\theta)}
{\mathrm{IR}^{\mathrm{raw}}_{q}(\theta_0)}.
\label{eq:ir-relative}
\end{equation}
Relative IR equals $1$ for the reference checkpoint. Values below $1$
indicate lower sensitivity than the reference, while values above $1$
indicate higher sensitivity. This comparison is defined within the same task,
training run, and model family; it does not make IR directly comparable
across them.

A checkpoint passes the screen only when relative IR is below its threshold
and SR is above its threshold. We combine these two requirements into one
score. Each term measures the checkpoint's normalized margin from one
threshold, and the minimum selects the weaker of the two results.

For model family $\mathcal F$ and visual shift $\upsilon$, let the thresholds
be $t_{\mathrm{IR}}$ and $t_{\mathrm{SR}}$. All measurements in a score use
the same shift. With a small numerical constant
$\varepsilon_{\mathrm{score}}$, define
\begin{equation}
S_{\mathcal F,\upsilon}(\theta;\theta_0)
=
\min\!\left\{
\frac{t_{\mathrm{IR}}-\mathrm{IR}^{\mathrm{rel}}_{q}(\theta;\theta_0)}
{|t_{\mathrm{IR}}|+\varepsilon_{\mathrm{score}}},
\frac{\mathrm{SR}_{q,\delta}(\theta)-t_{\mathrm{SR}}}
{|t_{\mathrm{SR}}|+\varepsilon_{\mathrm{score}}}
\right\}.
\label{eq:ir-sr-score}
\end{equation}
The score is nonnegative exactly when both requirements pass. A negative
score means that at least one requirement fails.

We choose the two thresholds using planning success on a set of source tasks
and apply them unchanged to the remaining tasks. Thresholds are selected
separately for each model family. After the thresholds are chosen, planning
success is no longer used to score a checkpoint. SR still uses the state
labels defined above.

To compare checkpoint $\theta_1$ with its reference $\theta_0$, we report the
change in screening score:
$\Delta S_{\mathcal F,\upsilon}(\theta_1;\theta_0)
=S_{\mathcal F,\upsilon}(\theta_1;\theta_0)
-S_{\mathcal F,\upsilon}(\theta_0;\theta_0)$.
A positive $\Delta S$ means that $\theta_1$ receives a better joint IR--SR
score than the reference. It does not label either checkpoint as robust.

The experiments evaluate the method at two levels. Pair-level experiments
test whether ACPC reflects changes in prediction error and planner selection
(\Cref{sec:exp-target-aligned,sec:exp-planner}). Checkpoint-level experiments
first select an IR--SR screen on some LeWM tasks and test it on the remaining
tasks under Gaussian noise. We then ask whether changes in the joint IR--SR
score agree with changes in planning success under blur and resize, and
whether PLDM shows the same low-IR, high-SR pattern
(\Cref{sec:exp-frozen-external,sec:exp-cross-stressor,sec:exp-pldm-diagnostics}).
These checkpoint-level findings are empirical; they do not follow from the
pairwise bounds.

\section{Experiments}\label{sec:exp}

After describing the evaluation protocol, we evaluate the diagnostic at two
levels. At the pair level, we test whether ACPC reflects prediction-error
change and the cost of a CEM plan change. At the checkpoint level, we examine
how IR and SR vary across LeWM recovery and whether a screen selected on some
tasks identifies recovery on the remaining tasks under Gaussian noise. We
then compare diagnostic and performance changes under blur and resize and
test whether the same low-IR, high-SR pattern appears on PLDM. We begin with a
local-geometry case study to visualize the representation pattern behind
these measurements.

\subsection{Evaluation Protocol}\label{sec:bg}

\paragraph{Models, Tasks, and Evaluation.}
Most experiments use LeWM~\cite{maes2026lewm}. We also evaluate
PLDM~\cite{sobal2025stresstesting}
to test the diagnostic on a second world-model architecture. We use four
control tasks: TwoRoom navigation, PushT planar manipulation, Reacher arm
control, and OGBench-Cube (Cube) 3D manipulation. For planning evaluation,
CEM uses the frozen world model to optimize five-step action sequences. We
report the \emph{planning success rate}, defined as the percentage of episodes
that reach the task goal.

We use \emph{visual perturbation} for one transformed history and
\emph{visual shift} for the distribution of such transformations. The main
experiments add Gaussian noise with $\sigma=0.08$ to the history observations
while keeping the goal image clean. Additional experiments use Gaussian blur
($k=15$) and resize (scale $0.25$).

For each task and model family, the Gaussian-noise sweep contains nine
training conditions: one without noise augmentation and eight with
full-sequence Gaussian-noise augmentation at
$\stdmax{}\in\{0.01,\ldots,0.08\}$. Both LeWM and PLDM have three independent
training runs for each task--condition pair (seeds 3072/3073/3074). Each
trained checkpoint is evaluated with seeds 42/43/44, using 100 episodes per
evaluation seed. We use \emph{task--run cell} for one task and one training
run.

The training run is the unit of replication. The three evaluation seeds
measure only within-run variability, so the reported means and dispersions
across runs are descriptive rather than population estimates.

\paragraph{Diagnostic Protocol.}
IR uses 100 logged anchors, five Gaussian-noise draws at the
evaluation severity $\sigma=0.08$ (blur and resize diagnostics instead apply
the corresponding shift), eight predicted steps,
uniform horizon weights, within-anchor draw averaging, and a q90 summary
across anchors. To compute SR, the protocol selects, for each anchor, a nearby
history with a different endpoint-state label and checks whether the two
rollouts remain farther apart than raw IR plus $0.10$.
Appendix~\ref{sec:appendix-A} gives the exact pairing,
normalization, and threshold rules. Checkpoint comparisons use relative IR
from \Cref{eq:ir-relative}.

\paragraph{Threshold Protocol.}
A checkpoint in the LeWM Gaussian-noise training sweep meets the
\emph{success-rate criterion} if it
recovers at least 80\% of the improvement from the unaugmented checkpoint to
the best checkpoint at evaluation noise $\sigma=0.08$, while losing no more
than five percentage points on clean observations. Both constants were fixed
a priori. We select $(t_{\mathrm{IR}},t_{\mathrm{SR}})$ on every nonempty proper
subset of the four tasks and apply the thresholds unchanged to the remaining tasks. The one-,
two-, and three-task selection settings produce $4+6+4=14$ directional
threshold-selection/test partitions. Metrics first average training runs
within each test task and then weight tasks equally; checkpoint rows are never
treated as independent replicates. Success rates are used to select and
evaluate thresholds, but they are not inputs to ACPC, IR, or SR.

For blur and resize, each task uses the thresholds selected under Gaussian
noise on the other three tasks. We compare 24 checkpoint pairs (four tasks
$\times$ three
runs $\times$ two shifts). Each pair contains the unaugmented checkpoint and
the checkpoint trained at $\stdmax{}=0.08$. A pair is positive when success
under the shift improves by at least five percentage points and clean success
decreases by no more than five points. We make no blur- or resize-specific
adjustment. This comparison tests whether the within-family IR--SR ordering
remains aligned with relative checkpoint performance under new visual shifts.

\subsection{Local Geometry of Perturbed Views}
\label{sec:exp-local-geometry}

We begin with a PushT case study showing how visual noise can make perturbed
views of one history overlap with other histories. The ratio
$r/\mathrm{NN}$ compares the spread of a history's perturbed views with its
distance to the nearest other clean history. The disjoint-ball fraction
measures how often neighboring perturbation clouds remain separate. These
measurements provide only a local geometric picture. Neighbors are selected
in the learned space, and each clean history is rolled forward under its own
recorded actions. Representation collapse can also make every radius small.
The measurements motivate the controlled ACPC comparison but do not replace
it.

Using a fixed PushT case study, we compare a matched pair of LeWM
checkpoints: one unaugmented and one trained with full-sequence Gaussian-noise
augmentation at $\stdmax{}=0.08$. We inspect both the encoder output and the
representation after eight autoregressive rollout steps. All ratios and
fractions use the original 192-dimensional projected latent space in which
the planner computes costs; formal definitions and sampling details are in
\Cref{sec:appendix-local-geometry}. The t-SNE
visualization~\cite{vandermaaten2008tsne} is qualitative only.

\begin{figure}[t]
\centering
\includegraphics[width=\linewidth]{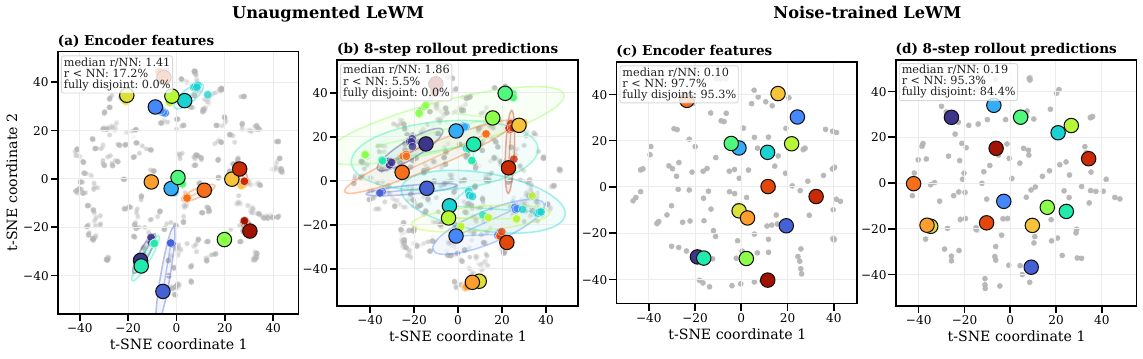}
\caption{Descriptive local-geometry case study motivating the controlled ACPC
comparison. Panels (a)--(d) compare a matched pair of PushT checkpoints---the
left two unaugmented, the right two trained with Gaussian-noise augmentation at
$\stdmax{}=0.08$---at the encoder output and after eight autoregressive
rollout steps. Each color denotes one of 16 highlighted
histories: the large black-rimmed marker is its clean anchor, smaller
same-color markers are its 18 perturbed views, and the faint ellipse is their
90\% t-SNE envelope; gray points show all other histories and views. Under
strong contraction, smaller markers and ellipses can lie beneath the clean
marker. Each panel's upper-left summary reports, across all 128 anchors, the
median $r/\mathrm{NN}$ and the percentages with $r<\mathrm{NN}$ and with fully
disjoint high-dimensional enclosing balls. The t-SNE coordinates and ellipses
are qualitative and enter none of these metrics.}
\label{fig:local-geometry-highd}
\end{figure}

\Cref{fig:local-geometry-highd} summarizes the case study.
Without augmentation, the median $r/\mathrm{NN}$ is $1.41$ at the encoder and
$1.86$ after eight rollout steps, and none of the 128 anchor clouds is fully
disjoint. After Gaussian-noise augmentation, the corresponding ratios fall to
$0.10$ and $0.19$; the fully disjoint fractions rise to $95.3\%$ at the
encoder and $84.4\%$ after the rollout. Thus, in this fixed case study,
augmentation makes perturbed views of the same history substantially more
compact relative to nearby clean histories. Much of this separation remains
after prediction.

\begin{figure}[t]
\centering
\includegraphics[width=\linewidth]{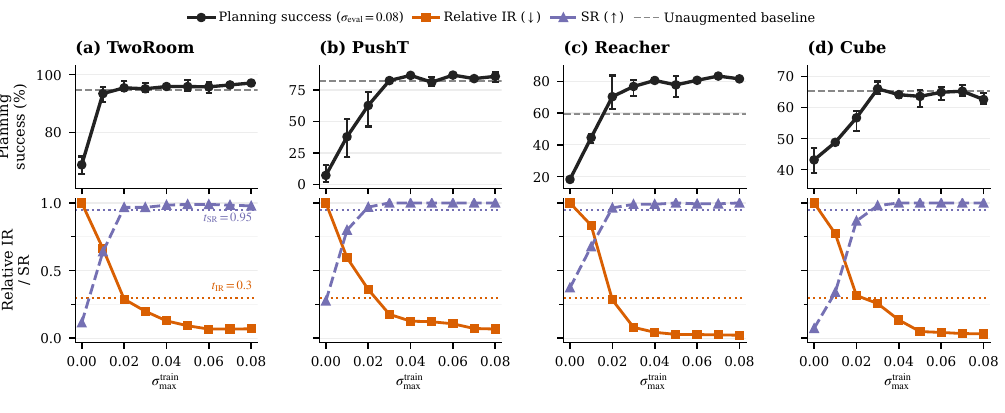}
\caption{Checkpoint-level planning performance and diagnostics across the
complete Gaussian-noise training sweep. We report planning success rate at
evaluation noise $\sigma=0.08$, relative IR, and SR. Points are across-run
means; success-rate error bars span the three training runs, and
success-rate axes are scaled per task. The dashed line is the clean success rate of the unaugmented
checkpoint; the leftmost black point is that checkpoint evaluated at
$\sigma=0.08$. Relative IR equals one there by construction. Lower IR and
higher SR are favorable. Dotted lines mark $t_{\mathrm{IR}}=0.3$ and $t_{\mathrm{SR}}=0.95$
(\Cref{sec:exp-frozen-external}).}
\label{fig:full-sweep-diagnostics}
\end{figure}

This case study shows that augmentation can contract perturbation-induced
variation relative to nearby clean histories, including after prediction. It
does not establish task-relevant separation or planning robustness. We
therefore turn to checkpoint-level IR and SR and their
relationship with planning performance across the complete augmentation
sweep (\Cref{fig:full-sweep-diagnostics}).

\subsection{IR and SR across Checkpoint Recovery}\label{sec:exp-gaussian-panel}

At evaluation noise $\sigma=0.08$, the unaugmented LeWM checkpoints lose
$25.9\pm2.4$, $74.6\pm5.6$, $41.0\pm1.4$, and $22.1\pm2.8$ percentage points
in success rate on TwoRoom, PushT, Reacher, and Cube, respectively.
Gaussian-noise augmentation recovers performance over a range of training
levels whose location differs by task. \Cref{fig:full-sweep-diagnostics}
compares success
rate with IR and SR across the complete sweep.

On Reacher, augmentation also raises clean success from $59.2\%$ to
$76$--$82\%$. Its recovery under noisy evaluation therefore cannot be
attributed to robustness alone; the checkpoints also improve on clean
observations.

Every augmentation level that meets the success-rate criterion has lower IR
and higher SR than its unaugmented reference, although neither measure is
monotone at every level. Across the 108 LeWM checkpoint rows, 77 pass the
reported IR threshold $t_{\mathrm{IR}}=0.3$, and all 77 also pass
$t_{\mathrm{SR}}=0.95$. Thus every accepted checkpoint combines reduced same-history sensitivity with retention
of the evaluated state-coordinate distinctions. The different-label distance
medians are largely stable across augmentation levels, so the rise in SR
mainly reflects contraction of the same-history radius rather than expansion
of different-label distances.

\paragraph{Evaluating SR under representation collapse.}
To test whether SR detects a loss of state separation, we train LeWM on
TwoRoom with four SIGReg weights, keeping all other settings fixed. Without
SIGReg, the representation collapses: the median latent distance shrinks from
$17.2$--$18.7$ to $0.006$, and clean success falls from $96.3$--$99.3\%$ to
$33.3\%$. Although this model has the lowest raw IR ($0.048$), its SR falls
to $0.066$, compared with $0.967$--$0.984$ for nonzero SIGReg. Thus, SR
exposes a loss of state separation that raw IR does not capture
(\Cref{tab:sigreg-collapse-control}).

IR tracks the recovery region, while SR verifies that the sampled
state-coordinate distinctions remain outside the contracted radius. A
local sensitivity analysis suggests that checkpoints with lower IR amplify
input perturbations less through the encoder and rollout
(\Cref{sec:appendix-local-sensitivity}). The next two experiments test whether
ACPC depends on the recorded actions and whether it reflects the cost of a
CEM plan change
(\Cref{sec:exp-target-aligned,sec:exp-planner}).
\FloatBarrier

\subsection{ACPC and Prediction-Error Change}\label{sec:exp-target-aligned}

Does ACPC help predict how much a visual perturbation changes multi-step
prediction error? For each history pair, \Cref{prop:target-free-error-drift}
shows that the two errors against the same logged future can differ by at most
$\mathrm{ACPC}_H$. We test whether measured ACPC is informative about this
bounded quantity, the error drift $d=|e_{\tilde h}-e_h|$. We evaluate at the protocol horizon
$H=8$, the longest horizon with a complete logged common future for every
anchor; \Cref{tab:horizon-quantile-sensitivity} shows that the
checkpoint-level conclusions are stable for $H\in\{1,2,4,8\}$.

\paragraph{Data and protocol.}
The analysis uses the unaugmented checkpoint of each task and training run.
Trajectories are partitioned into 16 disjoint groups. Each history pair
contributes rows at Gaussian-noise standard deviations
$\sigma\in\{0.02,0.05,0.08\}$ with two
perturbation draws; zero-severity probes serve only as identity checks. A
ridge model is fitted on 15 groups and evaluated on the remaining group,
rotating through all 16, and all rows from one trajectory group stay
together. We report the mean absolute error (MAE) on groups excluded from
model fitting.

\paragraph{Three nested regressions.}
Every model contains the probe severity and the encoder-history distance
$\|z-\tilde z\|_2$.
\begin{itemize}[leftmargin=1.4em]
  \item \textbf{Base} adds one-step ACPC under the recorded action: the
  information available without a multi-step rollout.
  \item \textbf{Base+Control$_8$} adds the strongest \emph{destroyed-action
  control}: eight-step ACPC recomputed with \emph{zeroed} actions (every
  action set to zero), \emph{swapped} actions (the recorded actions of a
  different trajectory), or \emph{shuffled} actions (the anchor's own actions
  in permuted temporal order). Each control performs the same eight-step
  rollout, so it carries rollout length without the recorded action sequence.
  For each task--run cell, we report the control with the lowest test MAE
  among the three, making this a conservative comparison.
  \item \textbf{Base+ACPC$_8$} instead adds eight-step ACPC under the
  recorded actions---the same actions associated with the observed future,
  and hence the feature matching the right-hand side of
  \Cref{prop:target-free-error-drift}.
\end{itemize}
All eight-step features use uniform weights $\alpha_k=1/8$ in
\Cref{eq:acpc-h}. If Base+ACPC$_8$ improves on Base, multi-step information
helps; if it also improves on Base+Control$_8$, the gain is attributable to
the recorded actions rather than to merely rolling out eight steps.

\begin{figure}[t]
\centering
\includegraphics[width=\linewidth]{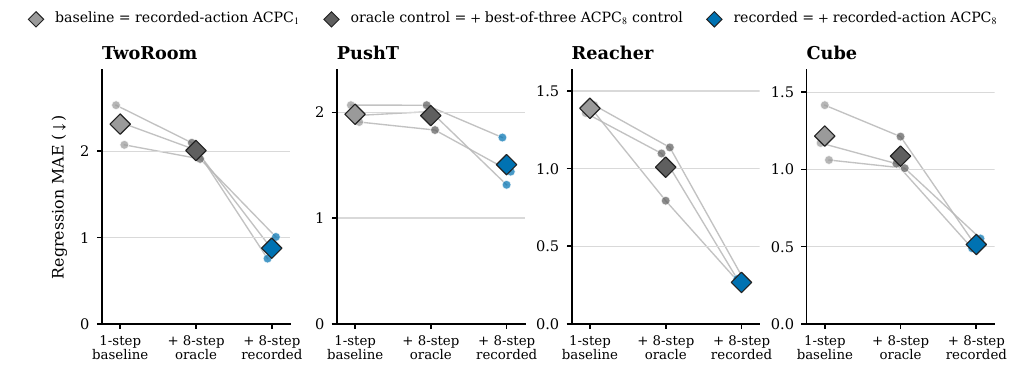}
\caption{Cross-validated MAE for predicting the error drift $d$ on the
unaugmented checkpoints; lower is better. Each model is evaluated on
trajectory groups excluded from fitting. Small points are training runs,
thin lines join the same run, and diamonds are task means. The in-figure
legend labels correspond to Base, Base+Control$_8$, and Base+ACPC$_8$ as
defined in \Cref{sec:exp-target-aligned}. Base+ACPC$_8$ is lowest in all 12
task--run cells; latent-error scales are task-specific, so compare within a
panel.}
\label{fig:future-drift-runs}
\end{figure}

\paragraph{Results.}
Base+ACPC$_8$ attains the lowest cross-validated MAE in all 12 task--run cells
(\Cref{fig:future-drift-runs}). For each training run we average the four
task-level relative MAE reductions equally and report the mean $\pm$ sample
standard deviation across the three training runs: the
reduction is $55.9\pm4.7\%$ relative to Base and $51.3\pm3.5\%$ relative to
Base+Control$_8$. These results concern prediction of the error drift; they
do not compare absolute prediction accuracy across rollout horizons.
\Cref{sec:appendix-p1-controls} reports the task-level values and selected controls
(\Cref{tab:target-aligned-acpc,tab:target-aligned-acpc-absolute}). Eight-step
ACPC carries information about the bounded error change that neither encoder
shift, one-step ACPC, nor any same-horizon destroyed-action control provides.

\subsection{ACPC and the Cost of CEM Plan Changes}\label{sec:exp-planner}

A visual perturbation may cause CEM to select a different plan. We ask whether
ACPC reflects how costly that change appears to the clean model.
Let $\mathbf a$ and $\tilde{\mathbf a}$ be the final plans selected from the
clean and perturbed histories, respectively. We measure
$(C_h(\tilde{\mathbf a})-C_h(\mathbf a))_+$: the extra clean-history model
cost incurred by selecting the perturbed-history plan. We refer to this
single-decision quantity as \emph{CEM selection regret}. It uses the model's
squared latent goal cost; it is neither cumulative regret nor simulator
return.

The fixed-pool bound motivates this comparison. However, adaptive CEM can
generate different later pools once the two elite sets diverge. We therefore
evaluate full paired CEM runs rather than treating the fixed-pool condition as
a run-level certificate.
The paired CEM branches follow \Cref{sec:appendix-planner-protocol}: same
initial proposal, common random numbers, each branch updating from its own
elites. For each candidate pool, we compute one- and five-step ACPC for every
action sequence and use the 90th percentile as the pool summary. The
five-step value matches the planner's prediction horizon. We compare two
ridge regressions. The baseline uses perturbation severity, training
condition, candidate-level one-step ACPC q90, and the clean gap between the
best and second-best candidates. The expanded model adds candidate-level
five-step ACPC q90. Within each training run, both regressions are fitted on
three tasks and tested on the remaining task, rotating which task is excluded
from fitting. We evaluate MAE on
$\log(1+\text{selection regret})$; Appendix~\ref{sec:appendix-planner-protocol}
gives the CEM budget and sample counts.

\begin{figure}[t]
\centering
\includegraphics[
  width=0.76\linewidth,
  trim=0 14pt 0 0,
  clip
]{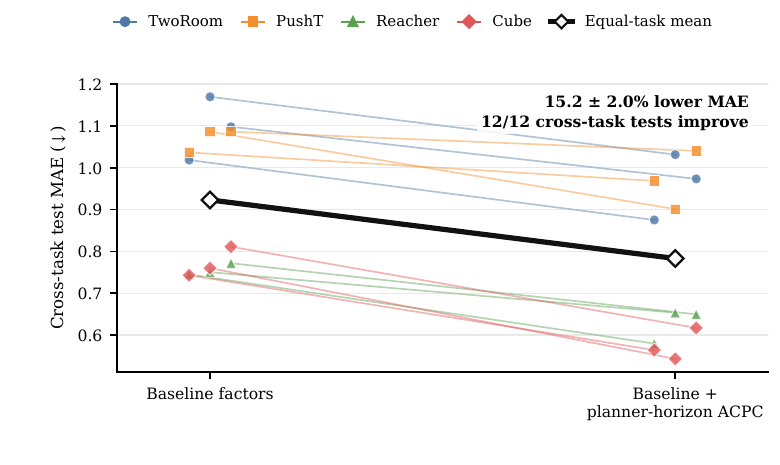}
\caption{Adding planner-horizon ACPC improves prediction of the extra model
cost associated with changes in the plan selected by adaptive CEM. Each
colored line connects the baseline and expanded-model errors for one test task
excluded from regression fitting in one training run; the thick black line is
the equal-task average across the three runs. The added feature is the 90th
percentile of candidate-specific ACPC at the planner's five-step prediction
horizon. The baseline already contains severity, training condition, the
corresponding one-step ACPC percentile, and the clean best--second-best cost
gap. MAE is computed on
$\log(1+\text{selection regret})$.}
\label{fig:acpc-planner-evidence}
\end{figure}

Adding planner-horizon ACPC lowers cross-task test MAE by
$15.2\pm2.0\%$ (mean $\pm$ sample standard deviation across training runs).
The error decreases in all 12 task--run test cases
(\Cref{fig:acpc-planner-evidence}). Since the two regressions differ only by
the planner-horizon feature, this gain shows that candidate-specific ACPC at
CEM's five-step horizon carries additional cross-task predictive information
beyond one-step ACPC and the clean cost gap. The prediction-error experiment
separately shows that recorded actions add information beyond same-horizon
controls (\Cref{sec:exp-target-aligned}).
Candidate-specific ACPC therefore helps predict the clean-model cost of a
perturbation-induced CEM selection change across tasks. The experiment does
not test whether using ACPC during planning improves task success.

\subsection{Checkpoint Screening across Tasks}
\label{sec:exp-frozen-external}

We test whether thresholds chosen on some tasks identify checkpoint recovery
on the remaining tasks. Because the training sweep is discrete and planning
success is estimated from a finite number of episodes, we do not attempt to
estimate an exact decision boundary. A decision is correct when it matches
the success-rate criterion of \Cref{sec:bg}. We also count the grid steps
between the first accepted checkpoint and the first checkpoint that meets
that criterion.

Across the 14 source/test splits, 13 choose
$(t_{\mathrm{IR}},t_{\mathrm{SR}})=(0.3,0.95)$; only the Reacher-only split
chooses $(0.1,0.95)$. With two or three source tasks, the first accepted checkpoint
is within $0.5$ grid steps of recovery on average and never differs by more
than one step (balanced accuracy $0.900$, precision/recall
$0.913/0.953$). Single-source selection is less reliable: balanced accuracy
averages $0.855$ and ranges from $0.723$ to $0.913$. Reacher alone chooses
the tighter IR threshold and delays acceptance on the other tasks by as many
as five levels (\Cref{sec:appendix-cross-task}).

The predominant rule accepts a checkpoint only when relative IR is at most
$0.3$ and SR is at least $0.95$. IR limits same-history sensitivity, while
SR requires the tested state-coordinate distinctions to remain separated.
Across the sweep, both measures improve over augmentation levels associated
with recovery (\Cref{fig:full-sweep-diagnostics}). Because $t_{\mathrm{IR}}=0.3$ is the
largest tested value, the upper edge of the useful range remains unresolved.
\FloatBarrier

\subsection{Diagnostic Behavior on PLDM}
\label{sec:exp-pldm-diagnostics}

We apply the same ACPC, IR, and SR definitions to PLDM, which uses a
different architecture and training recipe. The experiment repeats the
nine-checkpoint Gaussian-noise training sweep, paired histories, and
eight-step rollouts.

\begin{figure}[!htb]
\centering
\includegraphics[width=\linewidth]{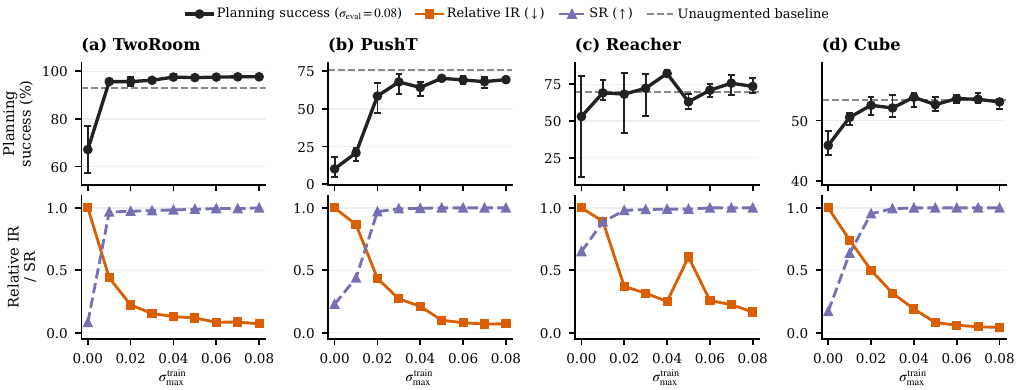}
\caption{PLDM Gaussian-noise training sweep over three independent training
runs. Curves show run means, and success-rate error bars span the run range. The
dashed gray line marks clean
success of the unaugmented checkpoint, and the leftmost black point is its
success at evaluation noise $\sigma=0.08$. Relative IR equals one at that
checkpoint. Across tasks, stronger augmentation generally lowers relative IR
and raises SR, while planning success under evaluation noise also generally
improves from the unaugmented checkpoint.}
\label{fig:pldm-sweep-diagnostics}
\end{figure}

Across all four PLDM tasks, the augmented checkpoints generally have lower
relative IR and higher SR than the unaugmented checkpoint. Planning success
under evaluation noise also generally improves from the unaugmented
checkpoint. These results reproduce the qualitative low-IR, high-SR pattern
under a second world-model architecture.

\subsection{Checkpoint Comparison under Blur and Resize}
\label{sec:exp-cross-stressor}

To test whether the diagnostics remain informative beyond Gaussian noise, we
evaluate one blur severity ($k=15$) and one resize severity (scale $0.25$).
For each task, training run, and visual shift, we compare the unaugmented
checkpoint with the checkpoint trained at $\stdmax{}=0.08$, giving
$4\times3\times2=24$ pairs. We recompute IR and SR under the evaluated shift
and use $\Delta S$ to compare the joint diagnostic scores of the two
checkpoints. A positive $\Delta S$ favors the augmented checkpoint. The
thresholds selected under Gaussian noise remain fixed, so this experiment
evaluates relative checkpoint ordering rather than an absolute pass decision.

\begin{figure}[!htb]
\centering
\includegraphics[width=\linewidth]{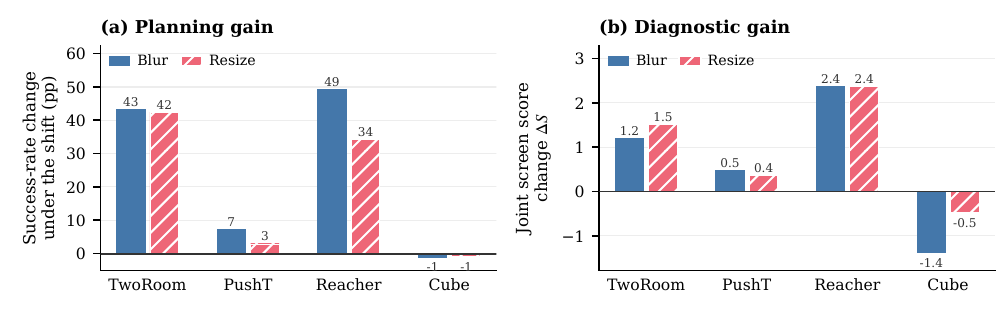}
\caption{Relative checkpoint comparison under blur and resize. Bars average
the three training runs for each task and shift; per-pair values are in
\Cref{tab:cross-stressor-ir-sr-all-pairs}. Panel (a) shows the change in planning
success from the unaugmented to the $\stdmax{}=0.08$ checkpoint. Panel (b)
shows the change in the IR--SR score of
\Cref{eq:ir-sr-score}; positive favors the augmented checkpoint
but is not an absolute pass decision.}
\label{fig:cross-stressor-ir-sr-comparison}
\end{figure}

In 22 of the 24 comparisons, the sign of $\Delta S$ agrees with whether the
augmented checkpoint meets the prespecified success criterion (balanced
accuracy $0.889$). Across all pairs, larger $\Delta S$ is associated with
larger gains in planning success (Spearman $\rho=0.835$). In the remaining
two cases, both $\Delta S$ and planning success improve, but the four-point
success gain falls just below the five-point criterion. We therefore treat
them as boundary cases rather than opposing diagnostic and behavioral trends.
\FloatBarrier

\section{Discussion and limitations}\label{sec:disc}

\paragraph{What is supported.}
For individual clean--perturbed pairs, multi-step ACPC captures changes in
prediction error more accurately than encoder distance, one-step ACPC, or
controls that remove the recorded action sequence. At the planning horizon,
ACPC also helps predict the extra model cost incurred when a perturbation
changes the plan selected by CEM, beyond one-step ACPC or the clean CEM
margin. Together, these results support the central claim behind our analysis:
rolling paired observations forward under the same action sequence reveals
downstream effects that encoder-only and one-step comparisons can miss.

Across checkpoints, planning success under Gaussian noise generally improves
as relative IR decreases and SR increases. Thresholds chosen using a subset of
tasks also identify recovery on tasks that were not used to choose them. Under
blur and resize, larger increases in the joint IR--SR score generally
correspond to larger gains in planning success. The fixed five-point criterion
produces only two boundary cases, each with a four-point success gain.

\paragraph{Why IR and SR are used together.}
IR can be low even when the representation collapses, because collapse brings
all predicted states closer together. SR prevents such a model from being
judged favorably by checking that states with different labels remain
distinguishable after rollout. SR alone is also insufficient because it does
not measure whether clean and perturbed observations remain close. IR
therefore measures sensitivity to visual perturbations, while SR measures
whether relevant state distinctions are preserved. The two measures should be
considered together.

\paragraph{Scope and limitations.}
ACPC diagnoses how visual perturbations affect predicted rollouts and planning
costs; it does not modify CEM. Our planner experiment therefore measures
whether ACPC reflects the extra predicted cost of a perturbation-induced plan
change, rather than whether ACPC-guided planning improves task success. The
adaptive-CEM guarantee applies only while its bound verifies that both runs
select the same elite candidates. The full IR--SR screen requires an
unaugmented reference checkpoint. Unlike pairwise ACPC, it also uses observed
future frames to normalize IR and dataset state labels to construct SR pairs.
Our checkpoint experiments cover Gaussian-noise augmentation and one severity
each of blur and resize. The chosen IR threshold is also at the edge of the
tested range, so larger values remain to be evaluated.

\paragraph{Future work.}
Future work can test whether using ACPC during planning improves task success.
Broader evaluation should include multiple perturbation severities, more
varied state pairs, additional robustness-training methods, and other JEPA
world models.

\section{Conclusion}\label{sec:conc}

ACPC measures how much a clean observation and its visually perturbed
counterpart diverge after both are rolled forward under the same action
sequence. We show that this divergence bounds changes in multi-step prediction
error and predicted planning costs. IR summarizes perturbation sensitivity
across a checkpoint, while SR checks whether different states remain
distinguishable after rollout. Experiments show that multi-step ACPC captures
changes in prediction error and the extra predicted cost of CEM plan changes
that encoder-only and one-step comparisons miss. Across checkpoints, lower IR
and higher SR generally align with better planning performance under visual
perturbations, and thresholds selected on some tasks identify recovery on the
remaining tasks. PLDM shows the same qualitative low-IR, high-SR pattern,
while the blur and resize results extend the evidence beyond Gaussian noise.
\bibliographystyle{unsrt}
\bibliography{references}

%% ===========================================================================
\FloatBarrier
\appendix
% TeX Live 2025's cleveref first aid records appendix sections as sections
% unless the section counter is explicitly aliased after \appendix.
\crefalias{section}{appendix}
\section{Proofs and Fixed-Pool Analysis}\label{sec:appendix-acpc-proofs}

\paragraph{Proof of \Cref{prop:target-free-error-drift}.}
Let $u=\bar G_{\mathbf a}(E_\theta(h))$,
$v=\bar G_{\mathbf a}(E_\theta(\tilde h))$, and $y=Y_{\mathbf a}^H$. Both
errors use the same target $y$ in the weighted rollout space of
\Cref{eq:weighted-rollout-map}. The reverse triangle inequality gives
\begin{align}
|e_{\tilde h}-e_h|
&=\bigl|\,\|v-y\|_2-\|u-y\|_2\,\bigr| \\
&\le \|(v-y)-(u-y)\|_2
=\|v-u\|_2
=\mathrm{ACPC}_H(h,\tilde h,\mathbf a).
\end{align}
The one-sided error increase satisfies the same bound because
$(e_{\tilde h}-e_h)_+\le|e_{\tilde h}-e_h|$.

The following result uses the cost-change bounds for individual candidates to
determine whether the selected candidate or elite set can change.

\begin{proposition}[Fixed-pool selection certificates]\label{prop:fixed-pool-certificates}
In the setting of \Cref{prop:exact-cost-certificate}, let $w$ be the unique
winner for the clean input. If its cost advantage over every other candidate
exceeds the combined cost-change bounds,
\begin{equation}
\min_{j\ne w}\bigl(C_j-C_w-b_j-b_w\bigr)>0,
\label{eq:top1-certificate-main}
\end{equation}
then it remains the winner for the perturbed input. Similarly, let $\mathcal
E$ be the clean top-$k$ elite set. If
\begin{equation}
\min_{i\in\mathcal E,\,j\notin\mathcal E}
\bigl(C_j-C_i-b_j-b_i\bigr)>0,
\label{eq:elite-certificate-main}
\end{equation}
then the perturbed input produces the same elite set.
\end{proposition}

\begin{corollary}[Conditional CEM stability]\label{cor:adaptive-cem-alignment}
Consider clean and perturbed CEM runs that start from the same proposal and use
the same random samples. Suppose that, at a given iteration, they evaluate
corresponding candidate action sequences. If
\Cref{eq:elite-certificate-main} holds, they select the same elite candidates
and fit the same proposal for the next iteration. If the condition holds at
every iteration, the two runs remain aligned. Because the evaluated solver
returns the final proposal mean as its action, the selected actions are also
identical.
\end{corollary}

\paragraph{Proof of \Cref{prop:exact-cost-certificate}.}
For a single candidate, omit the index. The difference between the two squared
costs factors as
\begin{align}
|\tilde C-C|
&=\bigl|\|\tilde x-g\|_2^2-\|x-g\|_2^2\bigr| \\
&=\bigl|\langle \tilde x-x,\tilde x+x-2g\rangle\bigr| \\
&\le \|\tilde x-x\|_2\,\|\tilde x+x-2g\|_2 \\
&\le r\bigl(\|x-g\|_2+\|\tilde x-g\|_2\bigr)=b .
\end{align}
The first inequality follows from Cauchy--Schwarz, and the second follows from
the triangle inequality.

Let $w$ and $\tilde w$ be the winners for the clean and perturbed inputs. Since
$\tilde w$ minimizes the perturbed cost,
\[
C_{\tilde w}
\leq \tilde C_{\tilde w}+b_{\tilde w}
\leq \tilde C_w+b_{\tilde w}
\leq C_w+b_w+b_{\tilde w},
\]
Since $w$ minimizes the clean cost, $C_{\tilde w}-C_w\geq0$. Together, these
inequalities prove \Cref{eq:fixed-pool-regret-bound}.

For any clean winner $w$ and competitor $j$,
\[
\tilde C_j-\tilde C_w
\ge C_j-C_w-|\tilde C_j-C_j|-|\tilde C_w-C_w|
\ge C_j-C_w-b_j-b_w.
\]
Therefore, \Cref{eq:top1-certificate-main} ensures that every competitor
remains more costly than $w$. Applying the same argument to every
$i\in\mathcal E$ and $j\notin\mathcal E$ proves that
\Cref{eq:elite-certificate-main} preserves the elite set. The strict
inequalities exclude ties and make the result independent of the tie-breaking
rule.

\paragraph{Proof of \Cref{cor:adaptive-cem-alignment}.}
Index the clean and perturbed runs by $b\in\{\mathrm c,\mathrm p\}$ and the CEM
iterations by $t$. Let $\phi_t^b$ denote the proposal mean and variance for
run $b$. Given shared random samples $\omega_t$, the deterministic sampling
map $T$ produces the candidate pool
\[
\mathcal A_t^b=T(\phi_t^b,\omega_t).
\]
Each run selects the indices $\mathcal E_t^b$ of its $k$ lowest-cost
candidates. A deterministic, permutation-invariant update $U$ then fits the
next proposal:
\[
\phi_{t+1}^b
=U(\{\mathbf a^j:j\in\mathcal E_t^b\}).
\]
The evaluated solver satisfies this condition because it uses the unweighted
mean and standard deviation of the elite candidates.

The two runs start from the same proposal, so
$\phi_0^{\mathrm c}=\phi_0^{\mathrm p}$. If their proposals agree at iteration
$t$, the shared samples generate the same candidate pool. When
\Cref{eq:elite-certificate-main} holds, the two runs also select the same elite
candidates and therefore fit the same next proposal:
\[
\phi_t^{\mathrm c}=\phi_t^{\mathrm p}
\;\Longrightarrow\;
\mathcal A_t^{\mathrm c}=\mathcal A_t^{\mathrm p}
\;\Longrightarrow\;
\mathcal E_t^{\mathrm c}=\mathcal E_t^{\mathrm p}
\;\Longrightarrow\;
\phi_{t+1}^{\mathrm c}=\phi_{t+1}^{\mathrm p}.
\]
Thus, if the condition holds at every iteration, the proposals and candidate
pools remain identical. The evaluated solver returns the final proposal mean,
so the selected actions are also identical.

If a solver instead returns the best candidate from the final pool,
\Cref{eq:top1-certificate-main} must also hold at the last iteration. If either
certificate cannot be verified, the proof no longer guarantees that the
subsequent elite sets, proposals, or candidate pools agree.

\paragraph{Relation to ACPC.}
The planning-cost bound uses the final-step displacement $r_j$, whereas ACPC
combines displacements across all $H$ rollout steps. From \Cref{eq:acpc-h},
\[
\sqrt{\alpha_H}\,r_j
\leq \mathrm{ACPC}_H
\qquad\text{when }\alpha_H>0.
\]
Thus, $r_j\leq\mathrm{ACPC}_H/\sqrt{\alpha_H}$. In the planner experiments, we
compute $r_j$ directly instead of using this upper bound. This gives a tighter
candidate-level measurement and does not require the observed future, but it
requires rolling out each candidate from both the clean and perturbed
histories.

\paragraph{Tightness of the bounds.}
Both bounds can hold with equality, so their right-hand sides cannot be
reduced without additional assumptions. For the prediction-error bound, let
$u$ be a unit vector, set the common target to $Y=0$, and let the two predicted
rollouts be $su$ and $tu$, where $s,t\geq0$. Then
\[
\bigl|\,\|tu\|_2-\|su\|_2\,\bigr|
=\|tu-su\|_2,
\]
so \Cref{eq:target-free-error-drift} holds with equality. For the planning-cost
bound, set $g=0$, $x=su$, and $\tilde x=tu$. Then
\[
|\tilde C-C|
=|t-s|(t+s)
=r\bigl(\|x-g\|_2+\|\tilde x-g\|_2\bigr),
\]
so the candidate cost-change bound is also attained exactly.

\section{Local-Geometry Case-Study Definitions}
\label{sec:appendix-local-geometry}

This section defines the high-dimensional geometric quantities reported in the
case study of \Cref{sec:exp-local-geometry}. We use 128 logged PushT histories.
For each history, we generate one clean view and 18 perturbed views, with six
draws at each standard deviation in $\{0.01,0.04,0.08\}$. Each of the four
qualitative t-SNE panels uses a separate deterministic projection, so
coordinates are not compared across panels.

We evaluate two representation spaces: the encoder output, denoted by
$\mathcal V=\mathrm{enc}$, and the representation after eight autoregressive
rollout steps, denoted by $\mathcal V=\mathrm{roll}$. The rollout horizon is
$H=8$ (\Cref{sec:acpc-def}). For history $i$, let
$v_{i,\mathcal V}^{(0)}$ be its clean representation and let
$v_{i,\mathcal V}^{(m)}$, $m=1,\ldots,M$ with $M=18$, be its perturbed
representations. In the rollout space, the clean and perturbed views of the
same history use the same recorded action sequence. We then define the sampled
visual radius, the distance to the nearest other clean history, and their ratio:
\begin{align}
r_{i,\mathcal V}^{\mathrm{vis}}
&=\max_{1\leq m\leq M}
\left\|v_{i,\mathcal V}^{(m)}-v_{i,\mathcal V}^{(0)}\right\|_2,
\nonumber\\
d_{i,\mathcal V}^{\mathrm{NN}}
&=\min_{j\neq i}
\left\|v_{i,\mathcal V}^{(0)}-v_{j,\mathcal V}^{(0)}\right\|_2,
\qquad
\rho_{i,\mathcal V}^{\mathrm{local}}
=\frac{r_{i,\mathcal V}^{\mathrm{vis}}}
{d_{i,\mathcal V}^{\mathrm{NN}}}.
\label{eq:local-radius-spacing}
\end{align}
The figures abbreviate $\rho_{i,\mathcal V}^{\mathrm{local}}$ as
$r/\mathrm{NN}$. A value below one means that every sampled perturbation moves
the representation by less than the distance from its clean anchor to any
other clean anchor. A value of at least one means that at least one
perturbation displacement is as large as the nearest-clean distance. This
comparison uses only distance magnitudes: it does not show that a perturbed
representation approaches or overlaps another history, and the nearest clean
history need not differ across a task-relevant state boundary.

The $r/\mathrm{NN}$ ratio considers only the perturbation radius of anchor
$i$. To account for the perturbation radii of both histories, we also compare
their enclosing balls. The ball around anchor $i$ is \emph{fully disjoint} if
it does not intersect the ball around any other anchor:
\begin{equation}
\left\|v_{i,\mathcal V}^{(0)}-v_{j,\mathcal V}^{(0)}\right\|_2
>
r_{i,\mathcal V}^{\mathrm{vis}}+r_{j,\mathcal V}^{\mathrm{vis}}
\quad\text{for every }j\neq i.
\label{eq:local-cloud-disjoint}
\end{equation}
Across the 128 anchors, we report three summaries: the median
$r/\mathrm{NN}$ ratio, the fraction with
$r_{i,\mathcal V}^{\mathrm{vis}}<d_{i,\mathcal V}^{\mathrm{NN}}$, and the
fraction satisfying the fully disjoint condition in
\Cref{eq:local-cloud-disjoint}. The second summary compares each perturbation
radius with the distance to the nearest other clean anchor. The third also
accounts for the perturbation radius around every other anchor. All three
summaries are computed in the original high-dimensional representation;
overlap between the qualitative t-SNE ellipses is not used.

These summaries are descriptive and depend on the sampled histories and
perturbation draws. They compare the encoder output and the final rollout step
but do not examine intermediate predictions. They also do not bound changes
in prediction error or planning cost; those theoretical links are provided by
pairwise ACPC in \Cref{sec:acpc-def}.

\section{Evaluation and Diagnostic Protocol}\label{sec:appendix-A}

This appendix specifies the fixed evaluation and diagnostic protocol used in
the main tables.

\paragraph{Evaluation grid.}
We evaluate LeWM on PushT, TwoRoom, Reacher, and Cube. Each checkpoint is
evaluated with seeds 42, 43, and 44, using 100 episodes per seed. The primary
evaluation adds Gaussian noise with standard deviation $0.08$ to the history
observations while keeping the goal image clean. For each task, the training
grid contains one unaugmented checkpoint and eight checkpoints trained with
full-sequence Gaussian-noise augmentation at
$\stdmax{}\in\{0.01,0.02,\ldots,0.08\}$.

\paragraph{PLDM protocol.}
PLDM is evaluated on the same four tasks and nine-checkpoint training grid as
LeWM, using the same evaluation seeds and number of logged histories. Its
diagnostics use the same clean--perturbed pairing, eight-step rollout horizon,
and SR definition. In the PLDM plots, raw IR is divided by the value of the
unaugmented PLDM checkpoint from the same task and training run.

\paragraph{Success-rate criterion.}
For each task and training run, let $P_{\mathrm{base}}$ be the
noisy-observation success rate of the unaugmented checkpoint, and let
$P_{\mathrm{best}}$ be the highest noisy-observation success rate in the
nine-checkpoint sweep. A checkpoint meets the success-rate criterion if its
noisy-observation success rate is at least
\[
P_{\mathrm{base}}+0.8(P_{\mathrm{best}}-P_{\mathrm{base}})
\]
and its clean-observation success rate is no more than five percentage points
below that of the unaugmented checkpoint. At the task level, we call an
augmentation level recovered only when this criterion holds in at least two
of the three training runs.

\paragraph{IR protocol.}
We compute IR separately for each task, training run, and checkpoint. To
express rollout divergence relative to the normal motion in each history, we
first compute a clean-motion scale. For anchor $i$, let
$u^{\mathrm{obs}}_{i,0}$ be the projected embedding of the final history
observation, and let $u^{\mathrm{obs}}_{i,1:H}$ be the projected embeddings of
the next $H$ observed clean frames. We define
\begin{equation}
s_i
=Q_{0.50}\!\left(
\left\{\|u^{\mathrm{obs}}_{i,k}-u^{\mathrm{obs}}_{i,k-1}\|_2
\right\}_{k=1}^{H}
\right).
\label{eq:clean-transition-scale}
\end{equation}
Thus, $s_i$ is the median displacement between consecutive clean observations,
including the transition from the final history observation to the first
future observation.

For each anchor, we generate multiple perturbation draws using the visual
shift being evaluated. The Gaussian-noise experiments use $\sigma=0.08$; the
blur and resize experiments apply their corresponding shifts. For every draw,
we compute eight-step ACPC with uniform weights $\alpha_k=1/H$ and normalize it
by $s_i+\varepsilon_{\mathrm{norm}}$. We average the normalized values across
draws for each anchor and then take q90 across anchors. The result is raw IR in
\Cref{eq:ir-raw}.

SR compares different-state distances with raw IR from the same checkpoint.
For the sweep plots, raw IR is divided by the value of the corresponding
unaugmented checkpoint, as defined in \Cref{eq:ir-relative}. Only the LeWM
cross-task analysis uses relative IR for threshold selection. We use
$\varepsilon_{\mathrm{norm}}=10^{-8}$ in
\Cref{eq:normalized-same-state-radius} and
$\varepsilon_{\mathrm{score}}=10^{-12}$ in \Cref{eq:ir-sr-score}.
\Cref{tab:horizon-quantile-sensitivity} reports the results for the tested
rollout horizons and summary quantiles.

\input{tables/table_horizon_quantile_sensitivity_ir_sr_v2}

\paragraph{SR protocol.}
SR follows \Cref{eq:sr}. We first define a local neighborhood using the
standardized logged endpoint states. Let $d_{0.35}$ be the 35th percentile of
all pairwise distances between different endpoints---that is, the q35 of all off-diagonal Euclidean distances.
This cutoff keeps the comparisons local while retaining an eligible neighbor for most anchors;
\Cref{tab:sr-labels} reports the resulting counts.

For each anchor $i$, we select the nearest history $j(i)$ that has a different
endpoint-state label and lies within distance $d_{0.35}$. If no such history
exists, the anchor is excluded from SR. We roll both histories forward under
the anchor's recorded action sequence $\mathbf a_i$. This sequence was not
generally executed by the neighboring history, but using it for both rollouts
prevents action differences from affecting the comparison. Their normalized
rollout distance is
\begin{equation}
D_i^{\mathrm{diff}}
=
\frac{
\left\|\bar G_{\mathbf a_i}(E_\theta(h_i))
-\bar G_{\mathbf a_i}(E_\theta(h_{j(i)}))\right\|_2
}{s_i+\varepsilon_{\mathrm{norm}}}.
\label{eq:different-state-distance}
\end{equation}
The pair counts as separated when $D_i^{\mathrm{diff}}$ is greater than raw
q90 IR plus the fixed margin $0.10$.

For pairing, the endpoint state is taken after the eight observed future steps
and therefore records the state reached under that trajectory's own actions.
The dataset preprocessor standardizes each state coordinate, and neighborhood
distances use the full standardized state vector. Endpoint-state labels are
constructed by median splits of the task-specific coordinates listed in
\Cref{tab:sr-labels}, using the fixed batch of 100 endpoints generated with
anchor seed 9101. Because the logged endpoints, labels, and neighborhoods are
fixed before checkpoint evaluation, the selected pairs and eligible-anchor
counts are identical across checkpoints and training runs.

\begin{table}[H]
\centering
\caption{Endpoint-state labels used to construct SR pairs. The state is taken
at the eighth observed future step, and its coordinates are standardized using
full-dataset statistics. Counts report eligible and skipped histories among
the fixed 100 anchors. These labels define operational partitions for the
diagnostic; they are not semantic or action-relevance annotations.}
\label{tab:sr-labels}
\footnotesize
\setlength{\tabcolsep}{3pt}
\begin{tabularx}{\linewidth}{l l X r}
\toprule
Task & Logged state field (dim.) & Label construction $\ell(y)$ & Eligible / skipped \\
\midrule
TwoRoom & \code{pos\_agent} (2) & Median split of the standardized agent $x$ coordinate (\code{pos\_agent[0:1]}). & 61 / 39 \\
PushT & \code{state} (7) & Three-bit label from median splits of standardized block $x$, block $y$, and block angle (\code{state[2:5]}). & 98 / 2 \\
Reacher & \code{observation} (6) & Three-bit label from median splits of the two standardized joint velocities and their Euclidean norm (\code{observation[4:6]}). & 100 / 0 \\
Cube & \code{observation} (28) & Three-bit label from median splits of the first two standardized coordinates and $\|r\|_2$, where $r=\bar o_{0:3}-\bar o_{25:28}$. & 100 / 0 \\
\bottomrule
\end{tabularx}
\end{table}

\paragraph{Evaluating SR under representation collapse.}
\label{sec:appendix-sigreg-collapse}
We evaluate whether SR detects representation collapse using a controlled
TwoRoom ablation. We train LeWM with SIGReg weights
$\{0,0.01,0.02,0.03\}$ while keeping the architecture, data, training seed,
and optimization settings fixed. We measure clean planning success over three
evaluation seeds. IR and SR use the same rollout construction and aggregation
as the main experiments, with Gaussian noise at $\sigma=0.005$.

\begin{table}[H]
\centering
\small
\setlength{\tabcolsep}{6pt}
\caption{\textbf{SR under representation collapse on TwoRoom.}
Clean success is the mean and standard deviation across evaluation seeds
42, 43, and 44, with 100 episodes per seed. IR and SR are computed from
100 logged histories, five perturbation draws, an eight-step rollout, and
Gaussian noise with $\sigma=0.005$. Median latent distance is the median
pairwise $\ell_2$ distance between clean representations at the eighth
observed future step. Removing SIGReg collapses latent spread and produces the
lowest raw IR, while SR and clean success fall sharply.}
\label{tab:sigreg-collapse-control}
\begin{tabular}{@{}ccccc@{}}
\toprule
SIGReg weight
& Clean success (\%)
& Median latent distance
& Raw IR
& SR \\
\midrule
$0$
& $33.3 \pm 3.7$
& $0.006$
& $0.048$
& $0.066$ \\
$0.01$
& $96.3 \pm 1.7$
& $17.174$
& $1.026$
& $0.967$ \\
$0.02$
& $99.3 \pm 0.5$
& $17.633$
& $0.144$
& $0.984$ \\
$0.03$
& $98.0 \pm 2.2$
& $18.686$
& $0.138$
& $0.984$ \\
\bottomrule
\end{tabular}
\end{table}

\paragraph{Threshold grids.}
For the LeWM cross-task evaluation, we search
\[
t_{\mathrm{IR}}\in\{0.05,0.075,0.10,0.15,0.20,0.30\},
\qquad
t_{\mathrm{SR}}\in\{0.80,0.85,0.90,0.95\}.
\]
For each nonempty subset of tasks used for threshold selection, we evaluate
every threshold pair. We first minimize the mean number of augmentation-grid
steps between the first checkpoint accepted by the IR--SR screen and the
first checkpoint that meets the success-rate criterion. Remaining ties are
resolved by, in order, fewer early acceptances, fewer late acceptances, higher
balanced accuracy, a smaller $t_{\mathrm{IR}}$, and a larger
$t_{\mathrm{SR}}$.

The selected thresholds are then applied unchanged to the tasks that were not
used for their selection. Success-rate labels from those tasks and all blur
or resize results are excluded from threshold selection.

\paragraph{Reproducibility.}
The commands used to rebuild the paper figures and tables are documented in
\codebrk{paper1/scripts/README.md} and
\codebrk{tools/README\_paper1.md}. The
released machine-readable summaries are sufficient for the reader-facing
aggregation and plotting steps. Recomputing checkpoint-level diagnostics
additionally requires the released checkpoints and datasets. The summaries
record the training seeds, evaluation seeds, and protocol hashes used for each
result.

\input{tables/table_full_sweep_compact_ir_sr_v2}

\FloatBarrier
\section{Prediction-Error Scope and Controls}\label{sec:appendix-p1-controls}

We checked \Cref{prop:target-free-error-drift} for every logged eight-step pair
and every candidate-specific five-step pair across all tasks, training runs,
and training conditions. No numerical violations occurred. This check
confirms that the implementation follows the inequality. It is not an independent statistical success count
or evidence of model quality, because the inequality holds for every correctly
computed pair.

The prediction-error and planning experiments answer different questions. The
prediction-error experiment measures how a visual perturbation changes
rollout error relative to the observed future under the recorded actions. The
planning experiment instead computes ACPC for the action candidates evaluated
by CEM and relates it to selection regret. We therefore treat the planning
analysis as separate evidence rather than infer it from the prediction-error
result (\Cref{sec:exp-planner}).

The prediction-error analysis uses only the unaugmented checkpoints and
Gaussian-noise levels $\sigma\in\{0.02,0.05,0.08\}$.
\Cref{tab:target-aligned-acpc} reports the relative reduction in
cross-validated MAE for each training run.
\Cref{tab:target-aligned-acpc-absolute} reports the corresponding MAE values
and the selected destroyed-action controls. Base+ACPC$_8$ achieves lower MAE
than both comparison models on all four tasks in every training run.

\input{tables/table_target_aligned_acpc}
\input{tables/table_target_aligned_acpc_absolute}

\FloatBarrier
\section{Cross-Task Threshold Partitions}\label{sec:appendix-cross-task}

With four tasks, there are 14 nonempty choices of tasks for threshold
selection: four single-task choices, six two-task choices, and four three-task
choices. The selected thresholds are applied to the remaining task or tasks.
Each evaluation includes all three training runs and all nine checkpoints for
every remaining task. Checkpoints are not counted as independent samples:
metrics first average the training runs within each evaluation task and then
weight the evaluation tasks equally.

When thresholds are selected using only Reacher, the procedure chooses
$(t_{\mathrm{IR}},t_{\mathrm{SR}})=(0.1,0.95)$ instead of the predominant
$(0.3,0.95)$. Both threshold pairs locate Reacher recovery within one
augmentation-grid step, but $t_{\mathrm{IR}}=0.1$ matches the Reacher onset
exactly and therefore wins the tie-break. Recovered Reacher checkpoints have
relative IR well below $0.1$, whereas the earliest recovered checkpoints on
the other tasks remain above $0.1$. The Reacher-only threshold therefore
accepts recovery on those tasks too late, by as many as five augmentation
levels.

\input{tables/table_cross_task_ir_sr_all_subsets_v2}

\FloatBarrier
\section{Cross-Stressor Pairs and Boundary Cases}
\label{sec:appendix-cross-stressor}

\input{tables/table_cross_stressor_ir_sr_summary_v2}

The remaining two cases lie at the boundary of the success-rate criterion:
PushT run 3072 under resize and PushT run 3074 under blur each improve success
rate by four percentage points, one point below the prespecified five-point
cutoff, while their IR--SR scores also increase. Thus, the diagnostic and
behavioral changes have the same direction; only the thresholded labels
differ. The complete table reports these cases alongside the aggregate
classification metrics.

\input{tables/table_cross_stressor_ir_sr_all_pairs_v2}

\FloatBarrier
\section{Local Sensitivity under Gaussian Noise}
\label{sec:appendix-local-sensitivity}

\begin{figure}[t]
\centering
\includegraphics[width=0.88\linewidth]{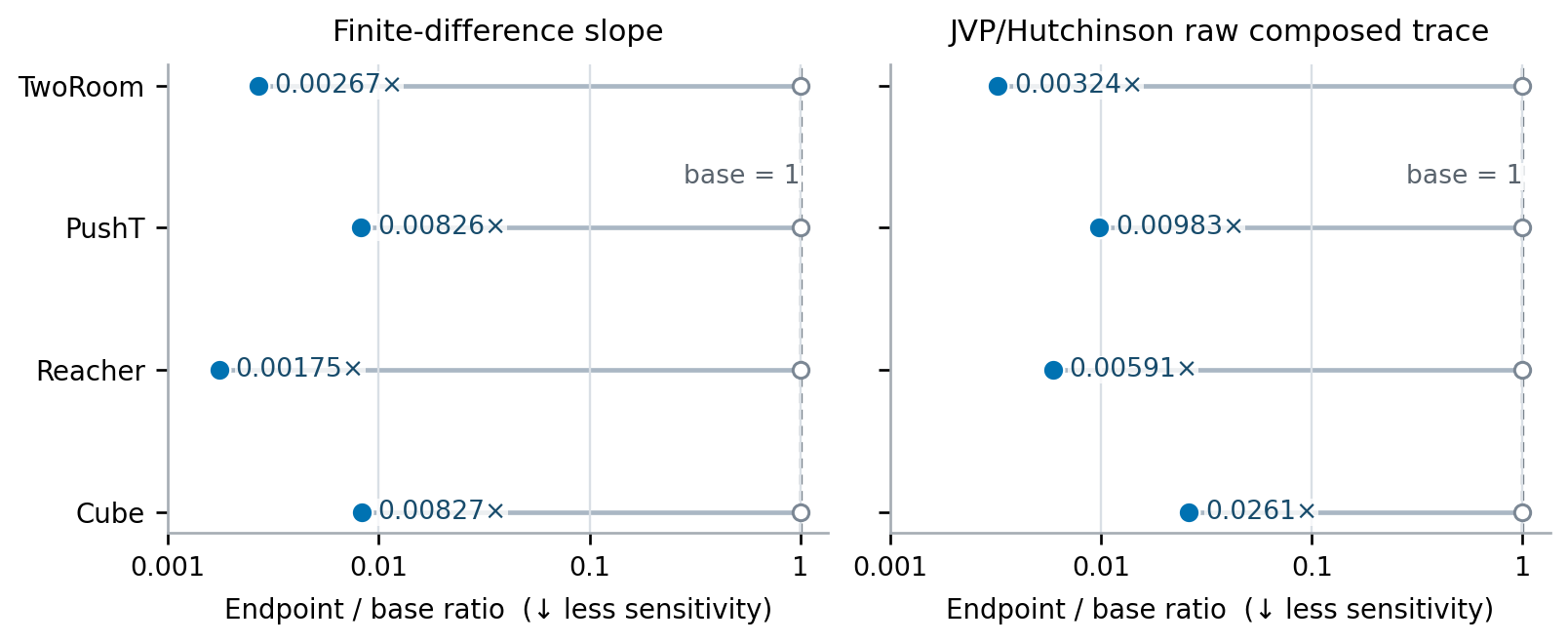}
\caption{Ratio of local sensitivity for the Gaussian-noise-augmented
checkpoint to that of the unaugmented checkpoint. We report a
finite-difference estimate and a JVP-based Hutchinson estimate of the composed
Jacobian's squared Frobenius norm. Both estimates are below one for every task,
indicating lower local sensitivity after augmentation. These local
measurements do not establish task performance or global robustness.}
\label{fig:local-gaussian-sensitivity}
\end{figure}

To relate ACPC under Gaussian noise to local model sensitivity, consider a
small isotropic perturbation $\delta x\sim\mathcal N(0,\sigma^2 I)$. Let $J_E$
and $J_G$ be the Jacobians of the encoder and rollout map, evaluated along the
clean input. A first-order expansion gives
\[
\Delta G=J_GJ_E\,\delta x+o(\|\delta x\|_2).
\]
Writing $A=J_GJ_E$ and using
$\mathbb E[\delta x\delta x^\top]=\sigma^2I$, we obtain
\begin{equation}
\mathbb E\!\left[\|A\delta x\|_2^2\right]
=\operatorname{tr}\!\left(
A\,\mathbb E[\delta x\delta x^\top]A^\top
\right)
=\sigma^2\operatorname{tr}(AA^\top)
=\sigma^2\|A\|_F^2,
\label{eq:local-gaussian-sensitivity}
\end{equation}
Therefore, for small $\sigma$,
\[
\mathbb E[\|\Delta G\|_2^2]
\approx \sigma^2\|J_GJ_E\|_F^2.
\]
The squared Frobenius norm of the composed Jacobian thus measures how strongly
the encoder--rollout path locally amplifies Gaussian observation noise.
Jacobian Frobenius norms have previously been used to measure and regularize
local representation sensitivity~\cite{rifai2011contractive}. We estimate
\[
\|J_GJ_E\|_F^2
=\operatorname{tr}\!\left[(J_GJ_E)^\top(J_GJ_E)\right]
\]
with Rademacher probes and the Hutchinson
estimator~\cite{hutchinson1989stochastic}. Jacobian--vector products compute
these estimates without constructing the full Jacobian. In every task, both
the finite-difference and Jacobian-based estimates are lower for the
noise-augmented checkpoint (\Cref{fig:local-gaussian-sensitivity}). This result
is consistent with Gaussian-noise augmentation reducing ACPC by lowering the
local sensitivity of the encoder--rollout path. The analysis applies only near
the evaluated inputs and does not provide a global robustness guarantee.

\FloatBarrier
\section{Adaptive-CEM Selection-Regret Protocol}\label{sec:appendix-planner-protocol}

The selection-regret analysis does not use task-success labels. We evaluate
the unaugmented and $\stdmax{}=0.08$ checkpoints from every task and training
run. For each checkpoint, we use 100 histories and apply perturbations at
severities $0.02$, $0.05$, and $0.08$. Each clean or perturbed CEM run uses 64
candidate action sequences, eight elite candidates, eight iterations, and a
five-step prediction horizon. The paired runs start from the same proposal and
use the same random samples, but each run updates its proposal using its own
elite candidates.

The implementation averages the squared embedding error across the $d$
embedding coordinates:
\[
C_j=\frac{\|x_j-g\|_2^2}{d}.
\]
This differs from the summed squared cost in
\Cref{prop:exact-cost-certificate} only by the positive factor $1/d$. The
factor rescales every $C_j$ and $b_j$ equally, so the regret bound and
selection certificates remain unchanged.

For each CEM run, we compute ACPC for every candidate under that candidate's
action sequence. We summarize the candidate pool by the q90 ACPC at horizons
one and five. Let $\mathbf a$ be the plan selected from the clean history and
$\tilde{\mathbf a}$ the plan selected from its perturbed counterpart. The
regression target is the extra predicted cost of the perturbed-history plan
when both plans are evaluated from the clean history:
\[
\bigl[C_h(\tilde{\mathbf a})-C_h(\mathbf a)\bigr]_+.
\]

Within each training run, we fit ridge regressions on three tasks and evaluate
them on the remaining task, rotating through all four choices. MAE is computed
on $\log(1+\text{selection regret})$ and averaged equally across the four
evaluation tasks. The relative MAE reduction is computed separately for each
training run before reporting the mean and sample standard deviation across
the three runs. This experiment evaluates whether ACPC reflects the model-cost
effect of a CEM selection change; it does not evaluate simulator return.

\end{document}

%% file: arxiv_metadata.tex
\newcommand{\arxivauthors}{%
Guo An$^{1,*}$,
Zijing Wu$^{2,*}$,
Honghua Dong$^{1,*}$\\
Yuhao Yan$^{3}$,
Zixuan Gui$^{4}$,
Haochong Chen$^{4}$,
Shanzhao Ruan$^{5}$\\
Xiang Wang$^{2}$,
Yurong Ling$^{6,\dagger}$,
Qi Tian$^{6,1,\dagger}$\\[0.35em]
{\small $^{1}$Huawei\quad
$^{2}$University of Science and Technology of China}\\
{\small $^{3}$Zhejiang University\quad
$^{4}$Tsinghua University\quad
$^{5}$Harbin Institute of Technology}\\
{\small $^{6}$Guangdong Laboratory of Artificial Intelligence and Digital Economy (SZ)}%
}
\newcommand{\arxivauthornotes}{%
$^{*}$Equal contribution.\quad
$^{\dagger}$Corresponding authors.\quad
Contact: Guo An, \texttt{anguo1@huawei.com}%
}
\newcommand{\paperpubliccodeurl}{https://github.com/Anguo-star/acpc-diagnostics}

%% file: tables/table_horizon_quantile_sensitivity_ir_sr_v2.tex
\begin{table}[H]
\centering
\caption{Horizon and quantile sensitivity of the IR reduction at fixed evaluation noise $\sigma=0.08$. Each entry is the median, over the three training runs, of the ratio between the IR of the checkpoint trained at the highest augmentation level ($\stdmax{}=0.08$) and that of the unaugmented checkpoint; lower means a larger reduction. These values are descriptive and do not retune any threshold.}
\label{tab:horizon-quantile-sensitivity}
\small
\setlength{\tabcolsep}{5pt}
\begin{tabular}{lccccccc}
\toprule
& \multicolumn{4}{c}{$q=0.90$ by horizon} & \multicolumn{3}{c}{$H=8$ by quantile} \\
\cmidrule(lr){2-5}\cmidrule(lr){6-8}
Task & H1 & H2 & H4 & H8 & q80 & q90 & q95 \\
\midrule
TwoRoom & 0.05 & 0.04 & 0.05 & 0.06 & 0.06 & 0.06 & 0.06 \\
PushT & 0.06 & 0.07 & 0.06 & 0.09 & 0.07 & 0.09 & 0.09 \\
Reacher & 0.02 & 0.03 & 0.03 & 0.02 & 0.02 & 0.02 & 0.02 \\
Cube & 0.04 & 0.03 & 0.04 & 0.04 & 0.03 & 0.04 & 0.04 \\
\bottomrule
\end{tabular}
\end{table}

%% file: tables/table_full_sweep_compact_ir_sr_v2.tex
\begin{table}[H]
\centering
\caption{Summary of the Gaussian-noise training sweep (nine checkpoints per task: no augmentation plus eight noise levels; \Cref{fig:full-sweep-diagnostics} shows every level). ``Best'' is the level with the highest mean planning success at evaluation noise $\sigma=0.08$; arrows give the change from the unaugmented checkpoint to that level. Relative IR is lower-is-better, and SR is higher-is-better. The last column lists levels meeting the success-rate criterion (\Cref{sec:bg}).}
\label{tab:full-sweep-compact}
\footnotesize
\setlength{\tabcolsep}{3.5pt}
\begin{tabular}{lrrrrrr}
\toprule
Task & No-aug. success (\%) & Best success (\%) & Best $\sigma_{\max}$ & Relative IR & SR & Levels meeting criterion \\
\midrule
TwoRoom & 68.8 & 97.1 & 0.08 & 1.00$\to$0.07 & 0.11$\to$0.98 & 0.01--0.08 \\
PushT & 7.2 & 86.8 & 0.06 & 1.00$\to$0.11 & 0.28$\to$1.00 & 0.03--0.08 \\
Reacher & 18.2 & 83.3 & 0.07 & 1.00$\to$0.03 & 0.37$\to$0.99 & 0.03--0.08 \\
Cube & 43.1 & 66.0 & 0.03 & 1.00$\to$0.26 & 0.07$\to$0.98 & 0.03--0.07 \\
\bottomrule
\end{tabular}
\end{table}

%% file: tables/table_target_aligned_acpc.tex
\begin{table}[!htb]
\centering
\caption{Relative reduction in out-of-group MAE from Base+ACPC$_8$ when predicting the error drift $d$ on the unaugmented checkpoints (protocol and model definitions in \Cref{sec:exp-target-aligned}). Base+Control$_8$ uses the per-cell oracle control; higher is better. The last column counts task--run cells in which Base+ACPC$_8$ is best.}
\label{tab:target-aligned-acpc}
\small
\setlength{\tabcolsep}{6pt}
\begin{tabular}{lrrc}
\toprule
Training run (seed) & vs. Base & vs. Base+Control$_8$ & Task--run cells improved \\
\midrule
3072 & 55.5\% & 51.4\% & 4/4 \\
3073 & 60.8\% & 54.8\% & 4/4 \\
3074 & 51.4\% & 47.7\% & 4/4 \\
\midrule
mean $\pm$ SD & 55.9 $\pm$ 4.7\% & 51.3 $\pm$ 3.5\% & 12/12 \\
\bottomrule
\end{tabular}
\end{table}

%% file: tables/table_target_aligned_acpc_absolute.tex
\begin{table}[!htb]
\centering
\caption{Out-of-group MAE (16-fold leave-one-trajectory-group-out) for predicting the error drift $d$ on the unaugmented checkpoints; lower is better, model definitions in \Cref{sec:exp-target-aligned}. ``Selected control'' is the destroyed-action control chosen by the per-cell oracle; ``paired wins'' counts the folds (of 16) in which Base+ACPC$_8$ beats Base+Control$_8$.}
\label{tab:target-aligned-acpc-absolute}
\footnotesize
\setlength{\tabcolsep}{3.2pt}
\begin{tabular}{lrrrrlc}
\toprule
Task & \shortstack{run\\(seed)} & Base & \shortstack{Base\\$+$Control$_8$} & \shortstack{Base\\$+$ACPC$_8$} & \shortstack{selected\\control} & \shortstack{paired\\wins} \\
\midrule
TwoRoom & 3072 & 2.535 & 2.099 & 0.755 & shuffled actions & 15/16 \\
TwoRoom & 3073 & 2.336 & 2.015 & 0.866 & shuffled actions & 15/16 \\
TwoRoom & 3074 & 2.075 & 1.911 & 1.006 & swapped actions & 13/16 \\
\addlinespace[1pt]
PushT & 3072 & 2.068 & 2.068 & 1.762 & shuffled actions & 11/16 \\
PushT & 3073 & 1.969 & 2.008 & 1.315 & zeroed actions & 13/16 \\
PushT & 3074 & 1.909 & 1.833 & 1.439 & zeroed actions & 13/16 \\
\addlinespace[1pt]
Reacher & 3072 & 1.358 & 1.097 & 0.288 & shuffled actions & 16/16 \\
Reacher & 3073 & 1.399 & 0.793 & 0.247 & shuffled actions & 16/16 \\
Reacher & 3074 & 1.409 & 1.136 & 0.263 & shuffled actions & 16/16 \\
\addlinespace[1pt]
Cube & 3072 & 1.171 & 1.037 & 0.489 & zeroed actions & 14/16 \\
Cube & 3073 & 1.416 & 1.212 & 0.500 & zeroed actions & 14/16 \\
Cube & 3074 & 1.061 & 1.008 & 0.552 & shuffled actions & 14/16 \\
\bottomrule
\end{tabular}
\end{table}

%% file: tables/table_cross_task_ir_sr_all_subsets_v2.tex
\begin{table*}[t]
\centering
\caption{Joint IR/SR checkpoint screening with thresholds chosen on other tasks. Thresholds are selected on the threshold-selection tasks and applied unchanged to all test tasks. Metrics first average training runs within each test task and then weight tasks equally. Recovery-onset mismatch is measured in training-augmentation grid levels (one level is $0.01$ in $\stdmax{}$).}
\label{tab:cross-task-all-subsets}
\small
\setlength{\tabcolsep}{3.5pt}
\begin{tabular}{>{\raggedright\arraybackslash}p{0.17\textwidth}>{\raggedright\arraybackslash}p{0.17\textwidth}rrrrr}
\toprule
Selection tasks & Test tasks & $t_{\mathrm{IR}}$ & $t_{\mathrm{SR}}$ & BA & P / R & \shortstack{Onset mismatch\\mean / max} \\
\midrule
TwoRoom & PushT, Reacher, Cube & 0.3 & 0.95 & 0.888 & 0.884 / 0.981 & 0.3 / 1.0 \\
PushT & TwoRoom, Reacher, Cube & 0.3 & 0.95 & 0.894 & 0.902 / 0.956 & 0.6 / 1.0 \\
Reacher & TwoRoom, PushT, Cube & 0.1 & 0.95 & 0.723 & 0.906 / 0.540 & 2.9 / 5.0 \\
Cube & TwoRoom, PushT, Reacher & 0.3 & 0.95 & 0.913 & 0.950 / 0.938 & 0.6 / 1.0 \\
TwoRoom, PushT & Reacher, Cube & 0.3 & 0.95 & 0.874 & 0.853 / 1.000 & 0.3 / 1.0 \\
TwoRoom, Reacher & PushT, Cube & 0.3 & 0.95 & 0.887 & 0.873 / 0.972 & 0.3 / 1.0 \\
TwoRoom, Cube & PushT, Reacher & 0.3 & 0.95 & 0.903 & 0.925 / 0.972 & 0.3 / 1.0 \\
PushT, Reacher & TwoRoom, Cube & 0.3 & 0.95 & 0.896 & 0.901 / 0.935 & 0.7 / 1.0 \\
PushT, Cube & TwoRoom, Reacher & 0.3 & 0.95 & 0.912 & 0.952 / 0.935 & 0.7 / 1.0 \\
Reacher, Cube & TwoRoom, PushT & 0.3 & 0.95 & 0.926 & 0.972 / 0.907 & 0.7 / 1.0 \\
TwoRoom, PushT, Reacher & Cube & 0.3 & 0.95 & 0.858 & 0.802 / 1.000 & 0.3 / 1.0 \\
TwoRoom, PushT, Cube & Reacher & 0.3 & 0.95 & 0.889 & 0.905 / 1.000 & 0.3 / 1.0 \\
TwoRoom, Reacher, Cube & PushT & 0.3 & 0.95 & 0.917 & 0.944 / 0.944 & 0.3 / 1.0 \\
PushT, Reacher, Cube & TwoRoom & 0.3 & 0.95 & 0.935 & 1.000 / 0.869 & 1.0 / 1.0 \\
\bottomrule
\end{tabular}
\end{table*}

%% file: tables/table_cross_stressor_ir_sr_summary_v2.tex
\begin{table}[!htb]
\centering
\caption{IR--SR scores under blur and resize. For each task, thresholds are chosen on the other three Gaussian-noise tasks and then fixed; each pair compares the $\stdmax{}=0.08$ checkpoint with its unaugmented counterpart. Balanced accuracy (BA), precision, and recall compare the sign of $\Delta S$ with the predefined five-point success criterion (\Cref{sec:exp-cross-stressor}). ``Boundary'' counts the two pairs with a four-point success gain and a positive $\Delta S$.}
\label{tab:cross-stressor-ir-sr-summary}
\small
\setlength{\tabcolsep}{4pt}
\begin{tabular}{lrrrrr}
\toprule
Visual shift & Pairs & BA & Precision / recall & Spearman & Boundary \\
\midrule
All pairs & 24 & 0.889 & 0.882 / 1.000 & 0.835 & 2 \\
Blur & 12 & 0.875 & 0.889 / 1.000 & 0.873 & 1 \\
Resize & 12 & 0.900 & 0.875 / 1.000 & 0.746 & 1 \\
\bottomrule
\end{tabular}
\end{table}

%% file: tables/table_cross_stressor_ir_sr_all_pairs_v2.tex
\begin{table*}[t]
\centering
\caption{All 24 LeWM checkpoint pairs evaluated under blur and resize. IR$_{\rm rel}$ and SR are measured for the checkpoint trained with Gaussian-noise augmentation ($\stdmax{}=0.08$); $\Delta P$ and $\Delta S$ are its success-rate and IR--SR score changes relative to the unaugmented checkpoint. The criterion is met when $\Delta P\geq5$ percentage points with at most a five-point clean loss. Daggers mark the two boundary cases with a four-point success gain, one point below the fixed cutoff, and a positive $\Delta S$.}
\label{tab:cross-stressor-ir-sr-all-pairs}
\footnotesize
\setlength{\tabcolsep}{3pt}
\begin{tabular}{lrlrrrrrl}
\toprule
Task & Seed & Shift & IR$_{\rm rel}$ & SR & $\Delta P$ & $\Delta S$ & Criterion / $\Delta S$ sign \\
\midrule
TwoRoom & 3072 & blur & 0.499 & 0.885 & 55.7 & 1.670 & met / positive \\
TwoRoom & 3072 & resize & 0.336 & 0.967 & 52.3 & 2.214 & met / positive \\
TwoRoom & 3073 & blur & 0.781 & 0.262 & 36.0 & 0.729 & met / positive \\
TwoRoom & 3073 & resize & 0.691 & 0.361 & 40.0 & 1.030 & met / positive \\
TwoRoom & 3074 & blur & 0.636 & 0.541 & 37.7 & 1.212 & met / positive \\
TwoRoom & 3074 & resize & 0.617 & 0.656 & 34.3 & 1.277 & met / positive \\
PushT & 3072 & blur & 0.943 & 0.939 & 10.7 & 0.189 & met / positive \\
PushT & 3072 & resize & 0.814 & 0.969 & 4.0 & 0.620 & not met / positive$^\dag$ \\
PushT & 3073 & blur & 0.846 & 0.918 & 7.3 & 0.515 & met / positive \\
PushT & 3073 & resize & 0.682 & 0.969 & 24.3 & 1.060 & met / positive \\
PushT & 3074 & blur & 0.781 & 0.959 & 4.0 & 0.729 & not met / positive$^\dag$ \\
PushT & 3074 & resize & 1.173 & 0.939 & -19.7 & -0.577 & not met / nonpositive \\
Reacher & 3072 & blur & 0.155 & 0.990 & 50.7 & 2.375 & met / positive \\
Reacher & 3072 & resize & 0.210 & 0.990 & 29.7 & 2.375 & met / positive \\
Reacher & 3073 & blur & 0.176 & 0.990 & 52.3 & 2.375 & met / positive \\
Reacher & 3073 & resize & 0.207 & 0.990 & 40.0 & 2.375 & met / positive \\
Reacher & 3074 & blur & 0.190 & 0.990 & 44.7 & 2.375 & met / positive \\
Reacher & 3074 & resize & 0.195 & 0.990 & 33.3 & 2.375 & met / positive \\
Cube & 3072 & blur & 1.447 & 0.330 & -2.7 & -1.488 & not met / nonpositive \\
Cube & 3072 & resize & 1.166 & 0.530 & 1.0 & -0.552 & not met / nonpositive \\
Cube & 3073 & blur & 1.577 & 0.260 & 2.3 & -1.923 & not met / nonpositive \\
Cube & 3073 & resize & 1.218 & 0.560 & -1.3 & -0.728 & not met / nonpositive \\
Cube & 3074 & blur & 1.220 & 0.660 & -3.0 & -0.735 & not met / nonpositive \\
Cube & 3074 & resize & 1.040 & 0.770 & -2.3 & -0.134 & not met / nonpositive \\
\bottomrule
\end{tabular}
\end{table*}